\PassOptionsToPackage{table}{xcolor}
\PassOptionsToPackage{hyphens}{url}
\documentclass[10pt,a4paper]{article}
\usepackage{zju-ant-preprint}
\usepackage[authoryear,round]{natbib}
\usepackage{bm,array,needspace,placeins,tabularx}

\usepackage{amsmath,amsfonts,bm}

\def\eqref#1{equation~\ref{#1}}
\def\1{\bm{1}}

\DeclareMathAlphabet{\mathsfit}{\encodingdefault}{\sfdefault}{m}{sl}
\SetMathAlphabet{\mathsfit}{bold}{\encodingdefault}{\sfdefault}{bx}{n}

\renewcommand{\bibfont}{\fontsize{8.8}{10.4}\selectfont}
\renewcommand{\PaperTitle}{One Evaluation Is Enough: Inversion-Free Image Editing with Trajectory Correction}
\renewcommand{\PaperShortTitle}{One Evaluation Is Enough}
\renewcommand{\PaperAuthors}{Lifan Jiang\textsuperscript{1,2},\quad Taiyang Chen\textsuperscript{3},\quad Senlin cheng\textsuperscript{2} \\[1pt]
  Bolei Xu\textsuperscript{2},\quad XIAOBO LI\textsuperscript{2},\quad Can Ye\textsuperscript{2} \\[1pt]
  Dehai Zhao\textsuperscript{1},\quad Deng Cai\textsuperscript{1},\quad Boxi Wu\textsuperscript{1,\textdagger}}
\renewcommand{\PaperAffiliations}{\textsuperscript{1}State Key Lab of CAD\&CG, Zhejiang University\qquad\textsuperscript{2}Ant Group\\[1pt]
  \textsuperscript{3}Chongqing Key Laboratory of Image Cognition,\\[1pt]
  Chongqing University of Posts and Telecommunications}
\renewcommand{\PaperContact}{lifanjiang@zju.edu.cn}
\renewcommand{\PaperAuthorNote}{\textsuperscript{\textdagger}Corresponding author: Boxi Wu.}
\renewcommand{\PaperKeywords}{Text-guided image editing; Inversion-free editing; One-step diffusion models; Endpoint-calibrated transport}

\renewcommand{\PaperDate}{October 2026}
\hypersetup{pdfauthor={Lifan Jiang, Taiyang Chen, Senlin cheng, Bolei Xu, XIAOBO LI, Can Ye, Dehai Zhao, Deng Cai, Boxi Wu}}
\renewcommand{\PaperAbstract}{%
Real-image editing commonly relies on inversion and iterative sampling, while efficient one-step editors often stabilize a large update by estimating and averaging editing fields at multiple noise levels. We ask a stricter question: is a single temporal field evaluation sufficient for reliable inversion-free editing? Our answer rests on a previously underused duality in a variance-preserving (VP) one-step model: one source-target query at a shared noisy state simultaneously determines a local angular semantic tangent and a clean endpoint displacement. We integrate the tangent over the analytically known remaining VP angle to obtain a nominal one-step transport, and use the endpoint displacement to correct its long-horizon extrapolation through a constrained projection. This transport-first, correction-second construction yields Endpoint-Calibrated Transport (ECT), a closed-form, training-free editor requiring only one noise level, two text conditions packed into one denoiser batch, and one latent update. ECT requires no image inversion, trajectory integration, second temporal observation, or iterative optimization, and can be directly applied to pretrained epsilon-prediction VP one-step generators. Across the reported benchmark settings, ECT delivers stable gains in editing quality within an extremely short editing time.
}

\begin{document}
\MakeJointTitle

\begin{figure*}[tp]
  \centering
  \includegraphics[width=\textwidth]{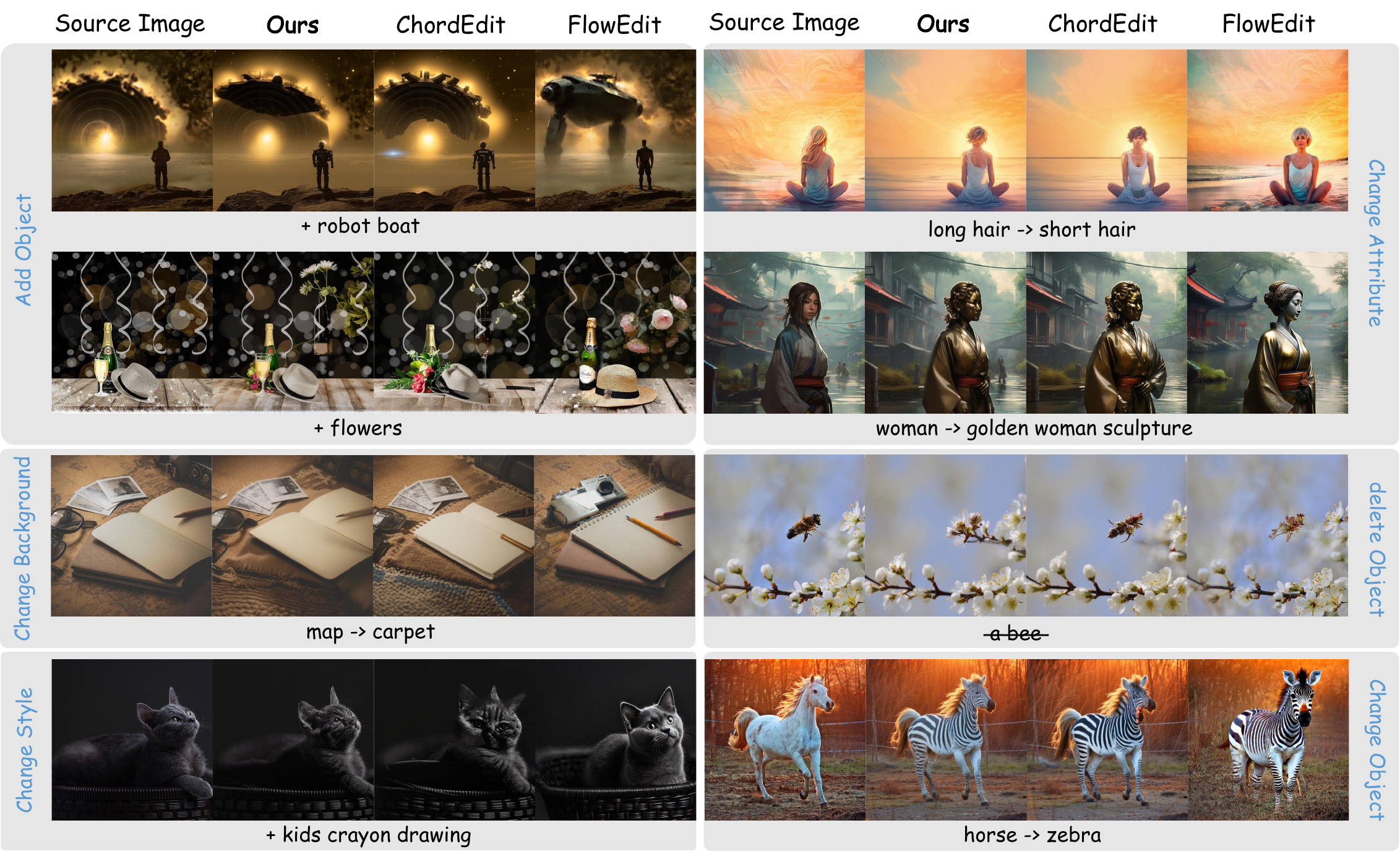}
  \caption{Edits with ECT (ours), ChordEdit, and FlowEdit, covering object addition and removal, attribute and background changes, stylization, and object replacement.}
  \label{fig:teaser}
\end{figure*}

\section{Introduction}
\label{sec:intro}

Text-guided real-image editing must accurately realize a target instruction while preserving unedited regions, spatial layout, and subject identity. Existing approaches use diffusion inversion, feature or attention injection, and iterative denoising to retain source structure. Although effective, inversion optimization and long sampling trajectories incur substantial latency and memory cost, limiting interactive use~\citep{hertz2023prompt,mokady2023null,tumanyan2023plug,cao2023masactrl}.

Recent work improves efficiency along two complementary directions. Inversion-free editors construct a direct or consistency-based process between source and target conditions, while distilled generators compress synthesis into one or a few steps~\citep{xu2024infedit,kulikov2025flowedit,liu2024instaflow,sauer2024add,dao2024swiftbrushv2}. Yet high-fidelity direct editors generally still integrate or correct model fields over multiple times. Even when the final edit uses one latent update, representative one-step methods stabilize that update with information collected at more than one noise level~\citep{lu2026chordedit}. Reliable editing from one temporal support thus remains challenging.

This paper studies \emph{single-evaluation image editing}: can a stable inversion-free edit be constructed by querying the source and target conditions at the same noisy state and at only one noise level? Multiple condition-time branches can be packed into a single batch, but every additional temporal observation still enlarges the batch, activation traffic, CPU scheduling, and memory pressure. We therefore characterize an editor by its temporal supports, condition-time branches, batched denoiser calls, and latent updates. We seek stable edits without additional temporal observations.

Our key observation is that a single conditional query of a VP one-step model determines one \emph{edit axis}: the angular tangent and clean-endpoint displacement derived from the shared state are collinear. We calibrate its magnitude by integrating the frozen local tangent over the known remaining VP angle, obtaining a nominal \emph{single-evaluation angular transport}. This schedule-based rescaling is a first-order approximation, not a guarantee that the original displacement underestimates the true edit. A closed-form endpoint-exposure shrinkage then moderates the extrapolation along the same axis, yielding \emph{Endpoint-Calibrated Transport} (ECT). The editor needs one noise level, two conditional branches in one denoiser batch, and one latent update, without training, inversion, a second temporal observation, or iterative optimization.

We validate ECT on two real-image editing benchmarks and two representative one-step backbones. The results support using one shared-state evaluation for both semantic transport and endpoint calibration, yielding competitive edits with lower overhead.

Our contributions are threefold:

\begin{itemize}
    \item \textbf{Single-evaluation formulation.} We formulate inversion-free editing under a strict single-temporal-support constraint: the source and target conditions are evaluated at one shared noisy state and followed by one latent update. It separates temporal support from batching and exposes the activation, memory, and scheduling costs of additional branches.
    \item \textbf{Transport-and-correction theory.} We derive a schedule-based nominal transport with a Lipschitz quadrature bound and a closed-form endpoint-exposure projection on the same edit axis. A shrinkage-risk analysis motivates correction under explicit assumptions.
    \item \textbf{Training-free core setting.} ECT operates directly on pretrained epsilon-prediction VP one-step backbones using one noise level, two text branches packed in one denoiser batch, and one latent update, without training, inversion, ODE integration, or iterative optimization. The pipeline combines competitive editing quality with lower execution overhead.
\end{itemize}

\begin{figure*}[tp]
  \centering
  \includegraphics[width=\textwidth]{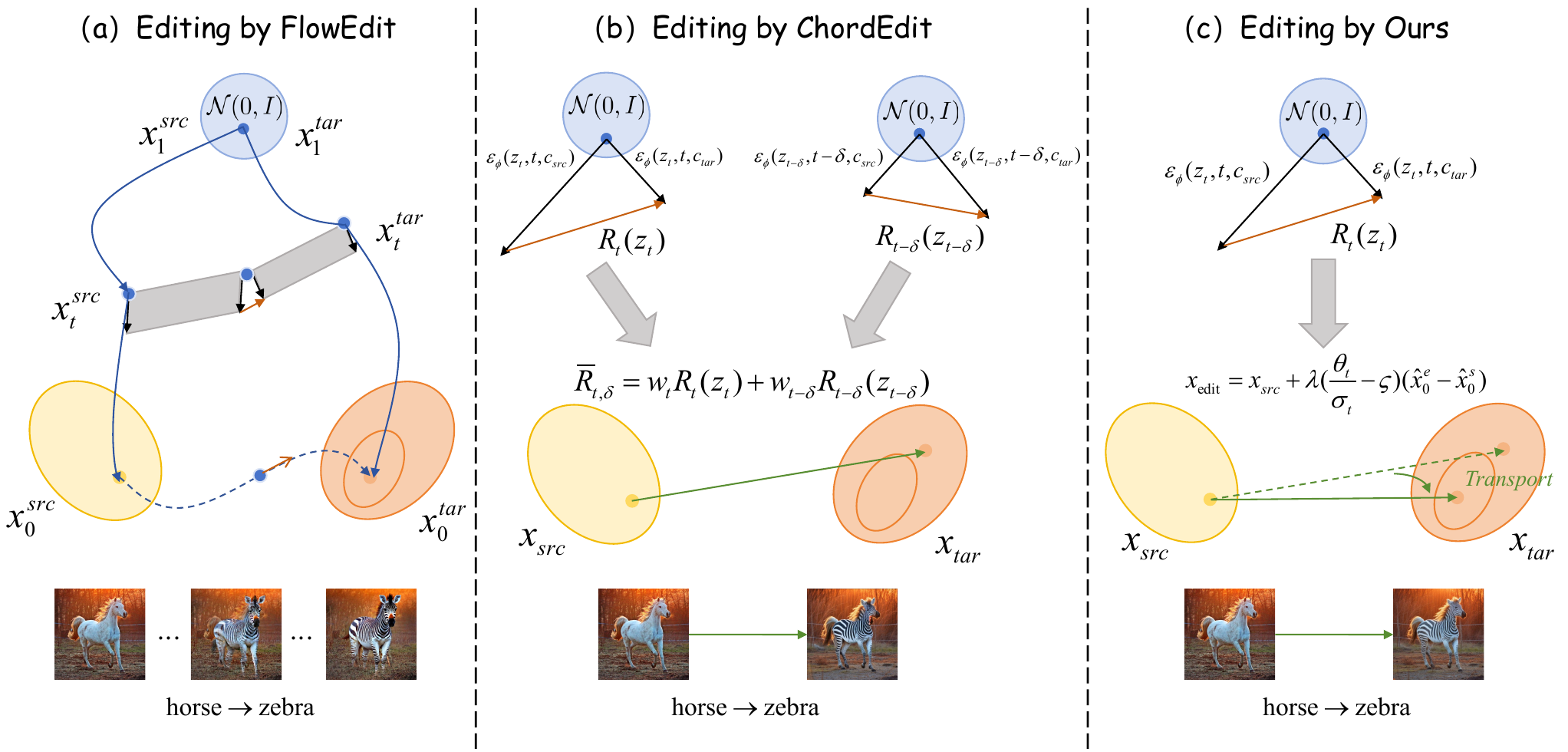}
  \caption{Editing pipelines, illustrated by a horse-to-zebra edit. (a) FlowEdit accumulates source--target field differences along a multi-step path. (b) ChordEdit combines residuals at $t$ and $t-\delta$ into one update, using two temporal supports and four condition--time branches. (c) ECT uses only $R_t(z_t)$ from two text conditions at one shared noisy state. The same predictions determine the angular residual $g_t$ and endpoint displacement $d_t$: angular transport gives $\widehat a_t=-\theta_t g_t$, and endpoint calibration gives $u_t^*=\widehat a_t-\zeta d_t$. A single update $x_{\mathrm{edit}}=x_{\mathrm{src}}+\lambda u_t^*$ then produces the edit (Equation~\ref{eq:closed_form}). The sketches are schematic; the update rescales a shared edit axis.}
  \label{fig:overview}
\end{figure*}

\section{Related Work}
\label{sec:related}

\textbf{Diffusion-based real-image editing.}
Diffusion probabilistic models and their deterministic samplers~\citep{ho2020ddpm,song2021ddim}, together with classifier-free guidance~\citep{ho2022cfg}, underlie most text-guided editors. Noise-and-denoise editing such as SDEdit~\citep{meng2022sdedit} and instruction-tuned editors such as InstructPix2Pix~\citep{brooks2023instructpix2pix} modify images directly in pixel or latent space, while text-guided diffusion editors manipulate cross-attention, inject intermediate features, or share self-attention to transfer target semantics while retaining source layout~\citep{hertz2023prompt,tumanyan2023plug,cao2023masactrl}. Their preservation ability commonly depends on DDIM inversion, null-text optimization, or source-reference branches that reconstruct an input-specific diffusion trajectory~\citep{mokady2023null,miyake2025negative}. Although these mechanisms provide fine-grained control, the reconstruction process introduces many sequential denoising steps, per-image optimization, and storage of intermediate activations or features. This creates a persistent trade-off between faithful reconstruction, editability, and interactive efficiency.

\textbf{Inversion-free direct editing.}
Inversion-free methods avoid reconstructing an exact noise code by coupling source and target predictions through consistency sampling or conditional transport~\citep{xu2024infedit,kulikov2025flowedit}. This direct view removes a major computational bottleneck and exposes the editing displacement itself as the central object of study. Trajectory-aware regularization can further preserve source structure by constraining conditional dynamics~\citep{kim2026flowalign}, and recent rectified-flow editors refine this direct transport through noise-domain alignment, structure-aware noise rectification, step-level inversion accuracy, velocity-discrepancy modeling, or frequency-interactive attention~\citep{xie2025dnaedit,jiang2026snredit,yang2026directedit,beaudouin2025drfs,yang2026fiaedit,xu2025ftedit}, but existing formulations generally observe, integrate, or correct the state along multiple ODE locations. We ask whether one temporal observation can provide both a global edit and its reliability correction.

\textbf{One-step generation and editing.}
Distillation, adversarial training, and rectified flow have compressed high-quality generation into one or a few steps~\citep{liu2023rectifiedflow,lipman2023flowmatching,song2023consistency,liu2024instaflow,sauer2024add,nguyen2024swiftbrush,dao2024swiftbrushv2}, building on large-scale latent and rectified-flow backbones~\citep{rombach2022ldm,podell2024sdxl,esser2024sd3,yan2024perflow,greenberg2025flux}. Editing is harder than unconditional generation because the single large update must simultaneously introduce target semantics and preserve input-specific content. Existing efficient editors therefore retain several generation steps or stabilize a final update through residuals evaluated at multiple times~\citep{deutch2024turboedit,gong2025instantedit,nguyen2025swiftedit,lu2026chordedit,li2026rethinking}. We use the VP angular horizon to obtain nominal global transport from one local residual, then calibrate it with the same query's clean endpoint signal.

\section{Method}
\label{sec:method}

ECT denotes the complete editor, including both analytic stages. We first convert a source-target conditional residual at one noise level into a nominal endpoint transport using VP angular geometry. We then apply a proximal endpoint correction to that transport under an endpoint-exposure constraint. Figure~\ref{fig:overview} contrasts this computation with multi-step FlowEdit and two-time ChordEdit, while Figure~\ref{fig:principle} illustrates the motivation for calibrating a single-step displacement.

\subsection{From Two-Time Averaging to One Evaluation}
\label{sec:problem}

Given a source image $I_{\mathrm{src}}$, source prompt $c_{\mathrm{src}}$, and target prompt $c_{\mathrm{tar}}$, a pretrained VAE encoder produces $x_{\mathrm{src}}=E(I_{\mathrm{src}})$. We seek a direct edit with one latent update,
\begin{equation}
    x_{\mathrm{edit}}=x_{\mathrm{src}}+u_t.
    \label{eq:direct_update}
\end{equation}
Here $u_t$ is the single latent edit vector, applied once at noise level $t$, that carries the source latent to its edited version; the whole method is a recipe for constructing it from one query. A residual observed at one time is local, and naively enlarging it into a global update can over-edit structure. A representative remedy averages residuals from two noise levels~\citep{lu2026chordedit}, but even batched into one call it still uses two temporal supports and four branches.

More generally, let the conditional editing residual at time $t$ be
\begin{equation}
    R_t(z_t)=\epsilon_\phi(z_t,t,c_{\mathrm{tar}})
    -\epsilon_\phi(z_t,t,c_{\mathrm{src}}).
    \label{eq:conditional_residual}
\end{equation}
The residual $R_t$ is the only condition-specific signal a single query provides, so every editor below---including ours---is ultimately a rule for turning $R_t$ into the update $u_t$ of Equation~\ref{eq:direct_update}. A two-time estimator constructs a stabilized direction in the generic form
\begin{equation}
    \bar R_{t,\delta}=w_tR_t(z_t)+w_{t-\delta}R_{t-\delta}(z_{t-\delta}),
    \qquad w_t+w_{t-\delta}=1,
    \label{eq:two_time_average}
\end{equation}
where the second observation reduces the uncertainty of extrapolating a local residual into one large update. Effective as it is, this couples editing to two schedule locations and doubles the condition-time width, even though the update is applied once.

\begin{figure}[tbp]
  \centering
  \includegraphics[width=\linewidth]{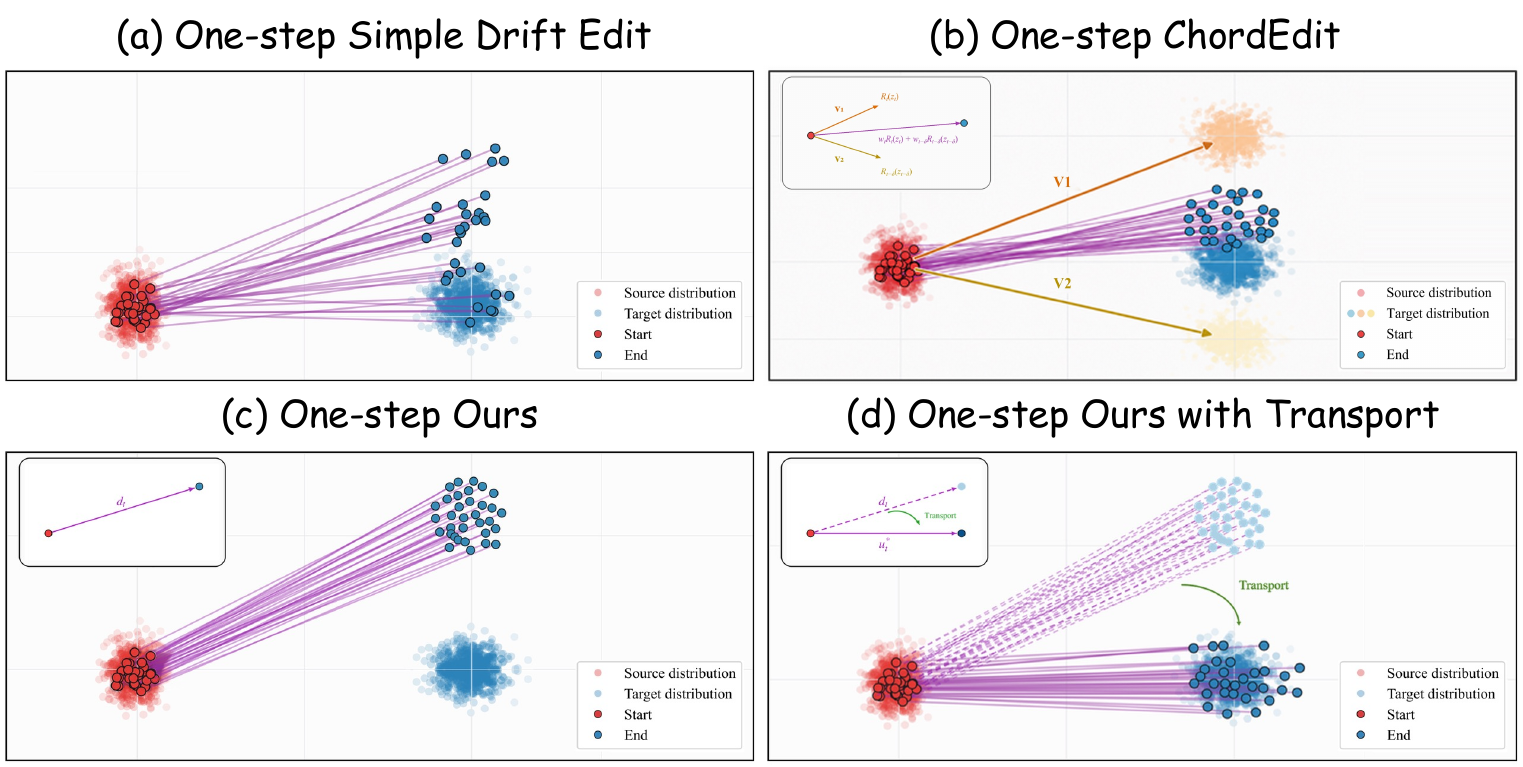}
  \caption{Schematic motivation for single-step calibration. (a) Simple drift produces dispersed endpoints. (b) ChordEdit combines two residuals. (c) Coherent updates can still miss the target distribution. (d) Transport and correction calibrate the endpoints. Direction changes abstract high-dimensional behavior; the exact ECT update is collinear with $d_t$ (Equation~\ref{eq:collinearity}).}
  \label{fig:principle}
\end{figure}

Figure~\ref{fig:principle}(a,b) illustrates the motivation: a direct single-step drift can produce dispersed endpoints, whereas combining two residuals can stabilize the final update. Panels~(c,d) illustrate the complementary need to calibrate a coherent but displaced set of endpoints. These are distribution-level cartoons, not measured trajectories or literal vector identities; our shared-state construction below performs the calibration by changing the magnitude along one edit axis.

We restrict $u_t$ to source-target predictions at one shared noisy state, formed by the VP process:
\begin{equation}
    z_t=\alpha_t x_{\mathrm{src}}+\sigma_t\epsilon,
    \qquad \epsilon\sim\mathcal{N}(0,I),
    \qquad \alpha_t^2+\sigma_t^2=1,
    \label{eq:vp_forward}
\end{equation}
we evaluate
\begin{equation}
    \hat\epsilon_s=\epsilon_\phi(z_t,t,c_{\mathrm{src}}),
    \qquad
    \hat\epsilon_e=\epsilon_\phi(z_t,t,c_{\mathrm{tar}}).
    \label{eq:single_query}
\end{equation}
Using the same $z_t$ and $t$ makes the prediction difference a comparison of text conditions at a fixed latent state. The common $z_t$ term cancels algebraically in the endpoint difference; independent noise draws would introduce a state difference as well. We convert this residual into a finite-horizon transport and calibrate its uncertainty through a deterministic map
\begin{equation}
    u_t=\mathcal{T}\!\left(
    z_t,t,\hat\epsilon_s,\hat\epsilon_e;\alpha_t,\sigma_t
    \right)
    \label{eq:single_support_map}
\end{equation}
that contains no model output from a second temporal support.

\subsection{Single-Evaluation Angular Transport}
\label{sec:angular}

To construct $\mathcal{T}$ in Equation~\ref{eq:single_support_map} from VP geometry, define
\begin{equation}
    \theta_t=\operatorname{atan2}(\sigma_t,\alpha_t),
    \qquad
    z_\theta=\cos\theta\,x_0+\sin\theta\,\epsilon.
    \label{eq:angular_parameterization}
\end{equation}
Differentiating this parameterization gives
\begin{equation}
    \frac{d z_\theta}{d\theta}
    =\cos\theta\,\epsilon-\sin\theta\,x_0.
    \label{eq:angular_derivative}
\end{equation}
The remaining angular interval to the clean endpoint is therefore $\theta_t$; this parameter interval need not equal the latent-space Euclidean arc length.
For condition $c$, one epsilon prediction at $z_t$ determines both its clean endpoint estimate and the corresponding angular tangent,
\begin{equation}
    \hat x_0^c=\frac{z_t-\sigma_t\hat\epsilon_c}{\alpha_t},
    \qquad
    \hat v_\theta^c=\alpha_t\hat\epsilon_c-\sigma_t\hat x_0^c.
    \label{eq:endpoint_velocity}
\end{equation}
The source-target angular residual is
\begin{equation}
    g_t=\hat v_\theta^e-\hat v_\theta^s.
    \label{eq:angular_residual}
\end{equation}
Here $g_t$ is the local angular-velocity change induced by replacing the source condition with the target. Let $g(\theta)$ denote this residual along the unknown editing trajectory. The ideal action is
\begin{equation}
    a_t^*=\int_{\theta_t}^{0}g(\theta)\,d\theta.
    \label{eq:ideal_transport}
\end{equation}
Although $g(\theta)$ away from $\theta_t$ is unobserved, its integration interval is fixed by the VP schedule. We therefore use the frozen-tangent approximation $g(\theta)\approx g_t$, giving the nominal transport
\begin{equation}
    \widehat a_t
    =\int_{\theta_t}^{0}g_t\,d\theta
    =-\theta_t g_t.
    \label{eq:nominal_transport}
\end{equation}
This uses the known angular horizon without querying a second temporal support. Section~\ref{sec:closed_form} gives the rescaling $\theta_t/\sigma_t\ge1$, and Appendix~\ref{app:angular_proof} explains its interpretation and limits.

We first bound the quadrature error for an exact local residual; model prediction error is separate.

\paragraph{Proposition 1 (single-evaluation nominal transport).}
Let $g(\theta)$ be the true source-target angular residual on $[0,\theta_t]$, assume it is $L$-Lipschitz, and take $g_t=g(\theta_t)$. Then
\begin{equation}
    \left\|a_t^*-\widehat a_t\right\|_2
    \le \frac{L}{2}\theta_t^2.
    \label{eq:quadrature_bound}
\end{equation}
The approximation is exact for a constant residual. A prediction error $\|g_t-g(\theta_t)\|_2\le\eta$ adds $\theta_t\eta$ to the bound (Appendix~\ref{app:angular_proof}); the bound does not determine its direction.

\subsection{Proximal Endpoint Correction}
\label{sec:correction}

Angular transport sets the nominal magnitude from schedule geometry by extrapolating one noisy estimate. Over a long horizon, the error in Proposition~1 can cause over- or under-shooting. We moderate the transport using the same predictions' clean endpoint displacement, requiring no extra evaluation:
\begin{equation}
    d_t=\hat x_0^e-\hat x_0^s.
    \label{eq:endpoint_displacement}
\end{equation}
We use $d_t$ as an endpoint-sensitive direction: $\langle d_t,u\rangle$ measures how strongly a candidate action advances along the endpoint displacement. Given an exposure budget $b_t$, ECT picks the action closest to the nominal transport while limiting this component,
\begin{equation}
    u_t^*=\arg\min_u\frac{1}{2}\|u-\widehat a_t\|_2^2
    \quad\text{s.t.}\quad
    \langle d_t,u\rangle\le b_t.
    \label{eq:ect_projection}
\end{equation}
This preserves the semantic direction from angular transport, changing it only as much as endpoint exposure requires. Its Lagrangian is
\begin{equation}
    \mathcal{L}(u,\zeta)
    =\frac{1}{2}\|u-\widehat a_t\|_2^2
    +\zeta\left(\langle d_t,u\rangle-b_t\right),
    \qquad \zeta\ge 0.
    \label{eq:ect_lagrangian}
\end{equation}
The associated KKT conditions are
\begin{equation}
\begin{aligned}
    u-\widehat a_t+\zeta d_t &=0, \\
    \langle d_t,u\rangle-b_t &\le 0, \\
    \zeta\big(\langle d_t,u\rangle-b_t\big)&=0.
\end{aligned}
    \label{eq:ect_kkt}
\end{equation}
Consequently, the unique closed-form correction is
\begin{equation}
    u_t^*=\widehat a_t-\zeta d_t
    =-\theta_t g_t-\zeta d_t.
    \label{eq:ect_update}
\end{equation}
If the nominal transport satisfies the constraint, $\zeta=0$ and it is unchanged; otherwise,
\begin{equation}
    \zeta=
    \frac{\langle d_t,\widehat a_t\rangle-b_t}{\|d_t\|_2^2},
    \qquad
    b_t=\langle d_t,\widehat a_t\rangle-\zeta\|d_t\|_2^2.
    \label{eq:dual_budget_relation}
\end{equation}
The implementation uses $\zeta$ directly, equivalent to the implicit budget in Equation~\ref{eq:dual_budget_relation}. This analytic shrinkage requires no further model evaluation. Appendix~\ref{app:risk} gives a conditional risk interpretation, rather than a guarantee for arbitrary model errors; Appendix~\ref{app:kkt} gives the projection derivation.

\subsection{Closed-Form Editor and Complexity}
\label{sec:closed_form}

Combining transport (Section~\ref{sec:angular}) and correction (Section~\ref{sec:correction}) gives a closed-form editor. The shared $z_t$ links $g_t$ and $d_t$ algebraically, eliminating numerical optimization. Equation~\ref{eq:endpoint_velocity} yields
\begin{equation}
    d_t=-\frac{\sigma_t}{\alpha_t}
    \left(\hat\epsilon_e-\hat\epsilon_s\right),
    \qquad
    g_t=\frac{1}{\alpha_t}
    \left(\hat\epsilon_e-\hat\epsilon_s\right).
    \label{eq:shared_state_residuals}
\end{equation}
Therefore,
\begin{equation}
    g_t=-\frac{1}{\sigma_t}d_t.
    \label{eq:collinearity}
\end{equation}
Substituting Equation~\ref{eq:collinearity} into Equation~\ref{eq:ect_update} yields the final editor,
\begin{equation}
    \boxed{
    x_{\mathrm{edit}}
    =x_{\mathrm{src}}
    +\lambda\left(\frac{\theta_t}{\sigma_t}-\zeta\right)
    (\hat x_0^e-\hat x_0^s)}.
    \label{eq:closed_form}
\end{equation}
Here $\lambda$ is the user-facing edit-strength coefficient, $\theta_t/\sigma_t$ the schedule-fixed transport gain, and $\zeta$ the endpoint-exposure shrinkage. The complete procedure is therefore
\begin{equation}
    (x_{\mathrm{src}},c_{\mathrm{src}},c_{\mathrm{tar}})
    \xrightarrow{\text{one }t}
    (\hat\epsilon_s,\hat\epsilon_e)
    \xrightarrow{\text{closed form}}
    x_{\mathrm{edit}},
    \label{eq:editor_pipeline}
\end{equation}
with no numerical solver between model prediction and latent update. Adding the endpoint difference to $x_{\mathrm{src}}$ also retains the source reconstruction residual $x_{\mathrm{src}}-\hat x_0^s$; replacing the source directly with $\hat x_0^e$ would discard this input-specific component.

\begin{figure*}[tp]
  \centering
  \includegraphics[width=\textwidth]{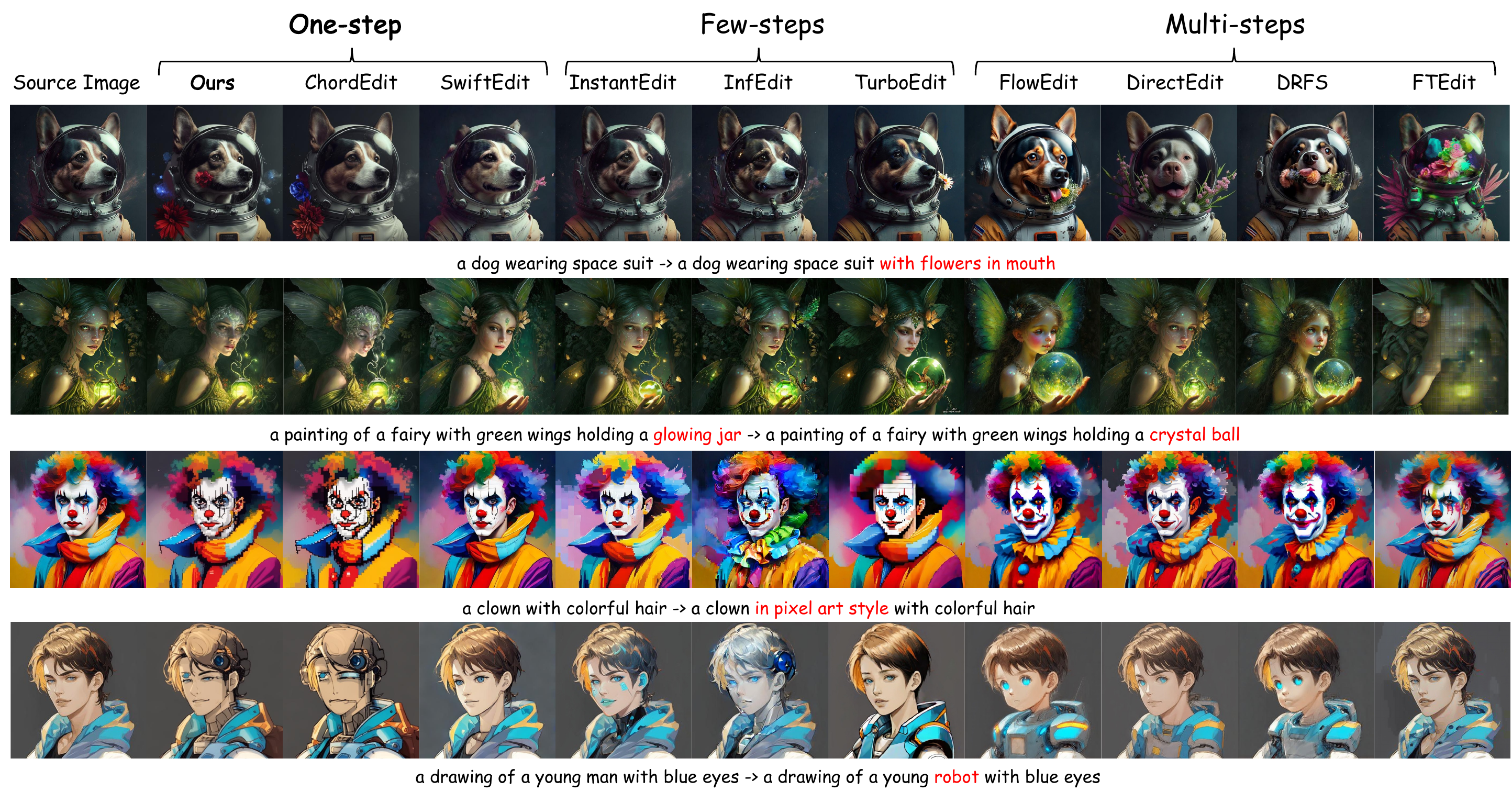}
  \caption{Qualitative comparisons of one-step, few-step, and multi-step editors. Rows show flower addition, crystal-ball replacement, pixel-art stylization, and robot transformation. Source images and instructions illustrate the preservation--alignment trade-offs.}
  \label{fig:qualitative}
\end{figure*}

ECT changes the magnitude along one edit axis. The nominal gain $\theta_t/\sigma_t\ge1$ follows from the frozen-tangent approximation, and $\zeta$ moderates that gain; neither operation supplies a new direction. With $\hat x_0^s\approx x_{\mathrm{src}}$, the update is approximately an interpolation toward the target endpoint when $0\le\lambda(\theta_t/\sigma_t-\zeta)\le1$. We evaluate this trade-off empirically.

The method constructs $z_t$ at one noise level, evaluates the two text conditions in one batch, and applies Equation~\ref{eq:closed_form} once, with neither inversion nor ODE integration. Section~\ref{sec:efficiency} reports temporal supports, condition-time branches, batched calls, latent updates, and measured efficiency.
\par\medskip

\section{Experiments}
\label{sec:experiments}
\subsection{Experimental Setup}
\label{sec:setup}
\textbf{Benchmarks and backbones.}
We evaluate on PIE-Bench and SNR-Bench with SD-Turbo (SD-T)~\citep{sauer2024add} and SwiftBrush-v2 (SB-v2)~\citep{dao2024swiftbrushv2}. The core uses one temporal support, two text-conditioned branches and one update, with $t=0.75$, $\zeta=0.7$, $\lambda=1$, and seed 42. Both main tables include unrefined ($0$) and refinement-enabled ($+$) ChordEdit/ECT; refinement adds one call at $t_{\rm prox}=0.3$. PIE additionally evaluates FLUX.1-Schnell (F-S)~\citep{bfl2024fluxschnell}, SD3.5~\citep{stabilityai2024sd35,esser2024sd3}, and native one-step image-to-image controls. Their linear flow paths admit a generalized VP representation through state normalization and angular reparameterization (Appendix~\ref{app:flow_derivation}). PIE contains 33 configurations covering 12 baseline algorithms and eight generator-checkpoint settings; SNR contains 21 configurations.

\begin{table*}[tp]
\caption{PIE-Bench results for 33 configurations. Adjacent ChordEdit/ECT rows share the same backbone and NFE budget. Bold denotes ECT values strictly better than ChordEdit at the displayed precision. CLIP-BPB uses common annotated backgrounds and follows Equation~\ref*{eq:bpb}.}
\label{tab:pie_results}
\centering
\setlength{\tabcolsep}{0.85pt}
\renewcommand{\arraystretch}{1.0}
\fontsize{9}{10}\selectfont
\begin{tabular*}{\textwidth}{@{\extracolsep{\fill}}lllrrrrrrrrrrrrr@{}}
\toprule
\multicolumn{3}{l}{Method / backbone} & \multicolumn{4}{c}{Background preservation} & Alignment & \multicolumn{4}{c}{Resources} & \multicolumn{3}{c}{Judge} & Rank \\
\cmidrule(lr){1-3}\cmidrule(lr){4-7}\cmidrule(lr){8-8}\cmidrule(lr){9-12}\cmidrule(lr){13-15}
Stage & Method & Model & PSNR $\uparrow$ & MSE $\downarrow$ & LPIPS $\downarrow$ & SSIM $\uparrow$ & CLIP-BPB $\uparrow$ & Time $\downarrow$ & VRAM $\downarrow$ & CPU $\downarrow$ & Mem. $\downarrow$ & CP $\uparrow$ & VQ $\uparrow$ & PC $\uparrow$ & Avg. $\downarrow$ \\
\midrule
M & DirectEdit & F-dev & 20.00 & 12.89 & 309.71 & 72.24 & 25.2903 & 14.690 & 37284 & 1024.80 & 6474.75 & 4.04 & 4.39 & 4.03 & 23.4167 \\
M & DNAEdit & F-dev & 21.85 & 8.34 & 219.16 & 78.41 & 25.7585 & 7.860 & 35881 & 1134.60 & 6468.61 & 4.02 & 4.45 & 4.11 & 19.4583 \\
M & SNR-Edit & F-dev & 21.08 & 9.75 & 204.70 & 77.66 & 25.8921 & 9.410 & 6446 & 2811.10 & 1333.25 & 4.13 & 4.59 & 4.32 & 14.6667 \\
M & FlowEdit & F-dev & 18.68 & 16.63 & 276.07 & 73.25 & 25.0768 & 8.280 & 37714 & 1148.80 & 6478.85 & 4.02 & 4.43 & 4.15 & 24.0417 \\
M & FlowEdit & SD3.5 & 19.17 & 14.51 & 267.66 & 73.38 & 26.9387 & 3.990 & 21873 & 1047.60 & 4737.02 & 3.97 & 4.29 & 3.88 & 22.7083 \\
M & DirectEdit & SD3.5 & 22.85 & 6.12 & 185.50 & 81.10 & 26.8927 & 5.400 & 21412 & 2536.40 & 4751.36 & 4.05 & 4.39 & 4.01 & 17.0417 \\
M & DNAEdit & SD3.5 & 22.93 & 6.20 & 199.40 & 80.82 & 26.8802 & 3.240 & 19924 & 1114.40 & 4721.66 & 4.02 & 4.39 & 4.02 & 16.5000 \\
M & DRFS & SD3.5 & 20.55 & 10.79 & 223.68 & 76.66 & 27.6072 & 5.420 & 19922 & 1233.70 & 4710.40 & 4.03 & 4.38 & 4.00 & 20.0000 \\
M & FIA-Edit & SD3.5 & 23.26 & 5.87 & 159.61 & 81.02 & 27.9635 & 5.330 & 19930 & 1321.90 & 4711.42 & 4.05 & 4.43 & 4.09 & 14.6667 \\
M & FTEdit & SD3.5 & 20.90 & 10.67 & 275.94 & 79.41 & 25.3104 & 31.240 & 26890 & 1379.90 & 2203.65 & 3.63 & 4.06 & 3.67 & 25.0833 \\
M & SNR-Edit & SD3.5 & 19.53 & 13.16 & 250.62 & 74.12 & 26.9664 & 5.250 & 6448 & 3390.90 & 1388.54 & 4.07 & 4.38 & 4.02 & 18.0417 \\
\midrule
F & TurboEdit & SDXL-T & 21.44 & 9.49 & 108.60 & 75.08 & 25.5718 & 0.610 & 14022 & 678.64 & 819.20 & 3.96 & 4.33 & 3.93 & 16.2500 \\
F & InfEdit & SD1.4 & 19.74 & 15.83 & 224.83 & 68.67 & 26.9412 & 0.481 & 7784 & 418.81 & 3364.86 & 3.85 & 4.27 & 3.83 & 21.4583 \\
F & InstantEdit & PeRFlow & 22.30 & 8.92 & 150.54 & 76.24 & 27.8933 & 0.670 & 9156 & 1293.70 & 843.78 & 4.03 & 4.39 & 4.03 & 14.3333 \\
\midrule
O & SwiftEdit & SB-v2 & 20.65 & 9.91 & 200.91 & 69.97 & 26.0762 & 0.451 & 16224 & 182.01 & 4282.37 & 4.07 & 4.53 & 4.22 & 15.3750 \\
O & ChordEdit$^{0}$ & SD-T & 23.89 & 6.46 & 88.36 & 81.26 & 24.6338 & 0.224 & 7492 & 408.14 & 1282.93 & 3.49 & 3.93 & 3.50 & 16.6250 \\
O & ECT$^{0}$ & SD-T & \textbf{26.00} & \textbf{4.11} & \textbf{64.12} & \textbf{83.93} & \textbf{25.4885} & \textbf{0.172} & \textbf{7478} & \textbf{206.42} & \textbf{1057.41} & \textbf{3.76} & \textbf{4.28} & \textbf{3.90} & \textbf{9.8750} \\
O & ChordEdit$^{+}$ & SD-T & 22.63 & 7.99 & 122.64 & 76.71 & 24.5982 & 0.265 & 7498 & 357.01 & 1297.07 & 3.95 & 4.37 & 4.00 & 15.4583 \\
O & ECT$^{+}$ & SD-T & \textbf{23.74} & \textbf{6.22} & \textbf{109.04} & \textbf{78.10} & \textbf{24.8320} & \textbf{0.212} & \textbf{7480} & \textbf{190.40} & \textbf{1075.96} & \textbf{4.09} & \textbf{4.65} & \textbf{4.31} & \textbf{8.4583} \\
O & ChordEdit$^{0}$ & SB-v2 & 20.81 & 10.78 & 197.23 & 70.33 & 24.9491 & 0.223 & 7492 & 414.37 & 1289.09 & 3.62 & 4.14 & 3.75 & 20.3333 \\
O & ECT$^{0}$ & SB-v2 & \textbf{22.85} & \textbf{6.73} & \textbf{141.96} & \textbf{74.65} & \textbf{25.6354} & \textbf{0.171} & \textbf{7478} & \textbf{191.67} & \textbf{1062.42} & \textbf{3.86} & \textbf{4.46} & \textbf{4.16} & \textbf{10.8333} \\
O & ChordEdit$^{+}$ & SB-v2 & 22.04 & 7.13 & 120.49 & 77.27 & 24.0644 & 0.265 & 7498 & 357.02 & 1298.74 & 3.91 & 4.37 & 4.05 & 15.1250 \\
O & ECT$^{+}$ & SB-v2 & \textbf{22.86} & 7.56 & 122.81 & 77.05 & \textbf{24.1936} & \textbf{0.213} & \textbf{7480} & \textbf{190.14} & \textbf{1075.93} & \textbf{4.03} & \textbf{4.45} & \textbf{4.13} & \textbf{11.0417} \\
O & ChordEdit$^{0}$ & F-S$^{\rm A}$ & 29.42 & 2.15 & 51.88 & 89.75 & 25.9676 & 0.587 & 32504 & 413.25 & 2926.28 & 3.66 & 4.14 & 3.69 & 16.0000 \\
O & ECT$^{0}$ & F-S$^{\rm A}$ & 29.09 & 2.38 & 55.01 & 89.30 & 25.7743 & \textbf{0.356} & \textbf{32338} & \textbf{203.80} & \textbf{2070.52} & \textbf{3.68} & \textbf{4.17} & \textbf{3.76} & \textbf{14.0000} \\
O & ChordEdit$^{+}$ & F-S$^{\rm A}$ & 25.98 & 4.22 & 75.80 & 82.77 & 25.4001 & 0.723 & 32469 & 347.88 & 2345.74 & 3.80 & 4.15 & 3.88 & 15.9583 \\
O & ECT$^{+}$ & F-S$^{\rm A}$ & 25.81 & \textbf{4.00} & 77.85 & 82.53 & 25.3436 & \textbf{0.498} & \textbf{32340} & \textbf{173.49} & 3036.17 & \textbf{3.83} & \textbf{4.25} & 3.83 & \textbf{15.5833} \\
O & ChordEdit$^{0}$ & SD3.5$^{\rm A}$ & 33.07 & 0.82 & 21.62 & 93.48 & 27.1649 & 0.242 & 15401 & 253.51 & 2618.16 & 3.54 & 3.98 & 3.51 & 13.5000 \\
O & ECT$^{0}$ & SD3.5$^{\rm A}$ & \textbf{34.57} & \textbf{0.66} & \textbf{19.44} & \textbf{93.79} & \textbf{27.1727} & \textbf{0.200} & \textbf{15385} & \textbf{129.34} & \textbf{2573.30} & \textbf{3.66} & \textbf{4.04} & \textbf{3.68} & \textbf{11.0417} \\
O & ChordEdit$^{+}$ & SD3.5$^{\rm A}$ & 25.87 & 4.27 & 129.64 & 79.56 & 23.1916 & 0.289 & 15393 & 241.18 & 2675.04 & 3.95 & 4.08 & 3.86 & 16.8333 \\
O & ECT$^{+}$ & SD3.5$^{\rm A}$ & \textbf{26.13} & \textbf{4.15} & \textbf{128.29} & \textbf{79.70} & 23.1299 & \textbf{0.248} & \textbf{15384} & \textbf{118.06} & 2756.88 & 3.90 & \textbf{4.12} & 3.86 & \textbf{15.2917} \\
O & Native 1-step & F-S$^{\rm A}$ & 11.81 & 91.64 & 385.71 & 54.64 & 17.3665 & 0.261 & 31464 & 524.08 & 1912.34 & 3.14 & 4.25 & 3.79 & 25.7917 \\
O & Native 1-step & SD3.5$^{\rm A}$ & 11.33 & 105.16 & 427.57 & 56.36 & 10.8362 & 0.213 & 15222 & 398.76 & 2545.79 & 2.50 & 3.12 & 2.44 & 26.2083 \\
\bottomrule
\end{tabular*}
\par\vspace{2pt}
\begin{minipage}{\textwidth}\fontsize{8}{9.5}\selectfont
$0$: no refinement; $+$: one extra call. M/F/O: multi-step, four-step and one-update editors. MSE/LPIPS $\times1000$; SSIM $\times100$; BPB is already on the CLIP-cosine $\times100$ scale. Time: s/image; VRAM/Mem.: MiB; CPU: core-\%. Avg.: mean rank across 12 criteria in this 33-row table. Full metric and cohort provenance is in Appendix~\ref*{app:provenance}.
\end{minipage}

\end{table*}

\begin{table*}[tp]
\caption{SNR-Bench results for 21 configurations. Adjacent ChordEdit/ECT rows share the same backbone and NFE budget. Bold denotes ECT values strictly better than ChordEdit at the displayed precision. CLIP-BPB uses the common automatic-region protocol in Appendix~\ref*{app:region_protocol}.}
\label{tab:snr_results}
\centering
\setlength{\tabcolsep}{0.85pt}
\renewcommand{\arraystretch}{1.0}
\fontsize{9}{10}\selectfont
\begin{tabular*}{\textwidth}{@{\extracolsep{\fill}}lllrrrrrrrrrrrrr@{}}
\toprule
\multicolumn{3}{l}{Method / backbone} & \multicolumn{4}{c}{Background preservation} & Alignment & \multicolumn{4}{c}{Resources} & \multicolumn{3}{c}{Judge} & Rank \\
\cmidrule(lr){1-3}\cmidrule(lr){4-7}\cmidrule(lr){8-8}\cmidrule(lr){9-12}\cmidrule(lr){13-15}
Stage & Method & Model & PSNR $\uparrow$ & MSE $\downarrow$ & LPIPS $\downarrow$ & SSIM $\uparrow$ & CLIP-BPB $\uparrow$ & Time $\downarrow$ & VRAM $\downarrow$ & CPU $\downarrow$ & Mem. $\downarrow$ & CP $\uparrow$ & VQ $\uparrow$ & PC $\uparrow$ & Avg. $\downarrow$ \\
\midrule
M & DirectEdit & F-dev & 15.96 & 42.66 & 450.31 & 51.22 & 24.9980 & 15.101 & 37284 & 102.81 & 6676.94 & 4.31 & 4.53 & 4.28 & 14.6667 \\
M & DNAEdit & F-dev & 19.52 & 15.05 & 250.75 & 64.82 & 25.9244 & 8.698 & 35881 & 103.86 & 6691.80 & 4.48 & 4.60 & 4.49 & 8.5000 \\
M & SNR-Edit & F-dev & 18.62 & 17.09 & 249.63 & 60.58 & 25.4541 & 9.454 & 6446 & 120.66 & 8267.65 & 4.46 & 4.58 & 4.34 & 9.6667 \\
M & FlowEdit & F-dev & 16.77 & 23.96 & 318.52 & 58.79 & 24.7119 & 8.653 & 37714 & 102.21 & 7246.46 & 4.41 & 4.56 & 4.38 & 12.0417 \\
M & FlowEdit & SD3.5 & 19.26 & 14.82 & 237.77 & 64.77 & 27.0241 & 3.112 & 21873 & 103.66 & 5242.98 & 4.45 & 4.59 & 4.44 & 7.6667 \\
M & DirectEdit & SD3.5 & 17.38 & 37.76 & 390.36 & 54.99 & 27.4479 & 5.147 & 21412 & 107.90 & 5144.69 & 4.28 & 4.45 & 4.23 & 13.0000 \\
M & DRFS & SD3.5 & 20.05 & 13.03 & 213.40 & 66.27 & 27.7279 & 5.360 & 19922 & 104.54 & 5270.13 & 4.50 & 4.69 & 4.48 & 6.4167 \\
M & FIA-Edit & SD3.5 & 17.86 & 26.76 & 332.32 & 56.92 & 28.3875 & 5.371 & 19930 & 112.90 & 5076.05 & 4.36 & 4.51 & 4.24 & 11.3750 \\
M & SNR-Edit & SD3.5 & 18.80 & 15.58 & 239.69 & 62.89 & 27.1593 & 5.350 & 6448 & 127.42 & 5097.49 & 4.34 & 4.56 & 4.43 & 8.0417 \\
\midrule
F & TurboEdit & SDXL-T & 14.77 & 40.80 & 430.38 & 46.42 & 25.0947 & 0.745 & 14022 & 141.18 & 1651.32 & 4.15 & 4.31 & 4.08 & 15.4167 \\
F & InfEdit & SD1.4 & 20.87 & 9.97 & 193.72 & 64.71 & 26.0500 & 0.434 & 7784 & 264.97 & 2777.93 & 4.23 & 4.50 & 4.28 & 8.2083 \\
F & InstantEdit & PeRFlow & 14.93 & 41.31 & 423.58 & 47.51 & 25.1356 & 0.815 & 9156 & 274.79 & 1804.33 & 4.10 & 4.36 & 4.05 & 16.0000 \\
\midrule
O & SwiftEdit & SB-v2 & 16.74 & 30.29 & 348.18 & 49.60 & 25.8191 & 0.472 & 16224 & 185.16 & 4756.48 & 4.26 & 4.44 & 4.28 & 13.5417 \\
O & ChordEdit$^{0}$ & SD-T & 18.95 & 15.90 & 280.64 & 61.04 & 23.3567 & 0.266 & 7492 & 379.08 & 1305.80 & 3.59 & 3.83 & 3.49 & 13.5417 \\
O & ECT$^{0}$ & SD-T & \textbf{20.91} & \textbf{9.97} & \textbf{203.62} & \textbf{66.31} & \textbf{24.5020} & \textbf{0.212} & \textbf{7478} & \textbf{189.47} & \textbf{1025.63} & \textbf{3.74} & \textbf{3.94} & \textbf{3.64} & \textbf{8.7500} \\
O & ChordEdit$^{+}$ & SD-T & 17.86 & 19.32 & 348.59 & 50.57 & 23.5146 & 0.305 & 7498 & 332.13 & 1316.07 & 4.14 & 4.46 & 4.15 & 13.9583 \\
O & ECT$^{+}$ & SD-T & \textbf{19.02} & \textbf{14.60} & \textbf{311.84} & \textbf{53.29} & \textbf{23.8777} & \textbf{0.257} & \textbf{7480} & \textbf{192.42} & \textbf{1035.02} & \textbf{4.30} & \textbf{4.48} & \textbf{4.30} & \textbf{9.3750} \\
O & ChordEdit$^{0}$ & SB-v2 & 20.12 & 11.88 & 222.36 & 64.68 & 23.7888 & 0.263 & 7492 & 385.18 & 1294.93 & 3.83 & 4.00 & 3.71 & 11.0417 \\
O & ECT$^{0}$ & SB-v2 & \textbf{21.87} & \textbf{7.83} & \textbf{165.57} & \textbf{68.94} & \textbf{24.6370} & \textbf{0.213} & \textbf{7478} & \textbf{183.42} & \textbf{1018.30} & \textbf{4.03} & \textbf{4.38} & \textbf{4.19} & \textbf{7.1250} \\
O & ChordEdit$^{+}$ & SB-v2 & 18.64 & 15.84 & 341.59 & 52.38 & 23.1999 & 0.305 & 7498 & 347.75 & 1312.53 & 4.21 & 4.41 & 4.21 & 13.3333 \\
O & ECT$^{+}$ & SB-v2 & \textbf{19.61} & \textbf{12.59} & \textbf{314.76} & \textbf{54.48} & 23.1162 & \textbf{0.254} & \textbf{7480} & \textbf{180.69} & \textbf{1037.11} & \textbf{4.30} & \textbf{4.44} & \textbf{4.30} & \textbf{9.3333} \\
\bottomrule
\end{tabular*}
\par\vspace{2pt}
\begin{minipage}{\textwidth}\fontsize{8}{9.5}\selectfont
$0$: no refinement; $+$: one extra call. M/F/O: multi-step, four-step and one-update editors. MSE/LPIPS $\times1000$; SSIM $\times100$; BPB is already on the CLIP-cosine $\times100$ scale. Time: s/image; VRAM/Mem.: MiB; CPU: core-\%. Avg.: mean rank across 12 criteria in this 21-row table. Full metric and cohort provenance is in Appendix~\ref*{app:provenance}.
\end{minipage}

\end{table*}

\textbf{Metrics.}
We report MSE, PSNR, SSIM~\citep{wang2004ssim}, and LPIPS~\citep{zhang2018lpips} on masked backgrounds to measure preservation, and CLIP-BPB for joint target-text alignment and background preservation. ImgEdit-Judge~\citep{ye2025imgedit} reports CP, VQ and PC on independent 1--5 scales. System efficiency uses synchronized per-image latency, time-weighted CPU-process utilization and RSS, and peak PyTorch-reserved VRAM. Appendices~\ref{app:image_metrics}--\ref{app:imgedit_judge} define them.

\textbf{CLIP-BPB.} Global CLIP scores the full output against the target text; its similarity can increase alongside changes to protected content. We therefore balance text compatibility with a background-preservation term, using fixed ViT-B/32 features~\citep{radford2021clip}:
\begin{equation}
\begin{aligned}
C_i&=100\cos\!\big(f_I(I_i^{\rm edit}),f_T(e_i)\big),\\
\operatorname{BPB}&=\frac{1}{|V|}\sum_{i\in V}C_i\exp(-1.8d_i),
\end{aligned}
\label{eq:bpb}
\end{equation}
where $d_i$ is unscaled SqueezeNet background LPIPS. The per-image product rewards outputs that combine target-text compatibility with source-background retention. PIE uses 556 common valid annotated backgrounds from 700 cases. SNR uses 80 cases with shared automatic regions and a fixed fallback, denoted BPB-Auto. We select the coefficient 1.8 from PIE method comparisons and keep it fixed on SNR; Appendix~\ref{app:region_protocol} reports the weight sweep and region protocol.

\textbf{Baselines.}
Our central comparison is the two-temporal-support ChordEdit estimator~\citep{lu2026chordedit} on the same one-step backbone. We include multi-step DirectEdit~\citep{yang2026directedit}, DNAEdit~\citep{xie2025dnaedit}, SNR-Edit~\citep{jiang2026snredit}, FlowEdit~\citep{kulikov2025flowedit}, DRFS~\citep{beaudouin2025drfs}, FIA-Edit~\citep{yang2026fiaedit}, and FTEdit~\citep{xu2025ftedit}; four-step TurboEdit~\citep{deutch2024turboedit}, InfEdit~\citep{xu2024infedit}, and InstantEdit~\citep{gong2025instantedit}; and one-step SwiftEdit~\citep{nguyen2025swiftedit} and ChordEdit. Native F-S/SD3.5 controls use one-step image-to-image generation.

\subsection{Main Quantitative Results}
\label{sec:main_results}
Tables~\ref{tab:pie_results} and~\ref{tab:snr_results} report quality and resource measurements at both refinement budgets. MSE/LPIPS are scaled by $1000$, SSIM by $100$, and BPB includes the cosine $\times100$ scale. Avg.\ rank averages all 12 criterion ranks within each table, using stored unrounded aggregates and average ranks for ties. Appendix~\ref{app:provenance} specifies the evaluation records and hardware groups.

ECT improves PSNR and all three judge scores on both original PIE backbones. SD-T also improves MSE, LPIPS and SSIM; refined SB-v2 has MSE $7.56$ versus $7.13$, LPIPS $122.81$ versus $120.49$, and SSIM $77.05$ versus $77.27$. Across both backbones and refinement budgets, ECT improves 45/48 quality/resource entries, including 24/24 without refinement and all four BPB comparisons ($+0.1292$ to $+0.8547$). Refined SD-T achieves the best PIE Avg.\ rank of $8.4583$.

On SNR, all four preservation metrics and all judge scores improve on both backbones, with BPB improvements in three of four matched pairs; refined SB-v2 scores $23.1162$ versus ChordEdit's $23.1999$. Unrefined SB-v2 gives ECT's best Avg.\ rank, $7.1250$, second among the 21 configurations. The logarithm in per-image PSNR permits higher mean PSNR alongside higher mean MSE. Matched backbone and refinement settings hold generator capacity and budget fixed. 

\textbf{Additional PIE backbones.} Unrefined SD3.5 increases PSNR from $33.07$ to $34.57$, reduces LPIPS from $21.62$ to $19.44$, and increases BPB from $27.1649$ to $27.1727$. On F-S, ECT reduces unrefined latency from $0.587$ to $0.356$ s and refined latency from $0.723$ to $0.498$ s; the corresponding BPB values are $25.7743/25.3436$ for ECT and $25.9676/25.4001$ for ChordEdit. ECT reduces latency on both flow backbones, with backbone-dependent preservation scores.

\subsection{System Efficiency}
\label{sec:efficiency}
Compared with matched ChordEdit rows on PIE, ECT reduces synchronized latency by $19.62$--$23.32\%$, CPU-process utilization by $46.67$--$53.74\%$, and host RSS by $17.05$--$17.58\%$. Peak reserved VRAM decreases by $0.19$--$0.24\%$. Savings are computed within the same device, backbone, and refinement budget. VAE and other shared work contribute to total latency; reducing condition--time branches lowers the denoiser and activation costs. Appendix~\ref{app:resources} defines the resource windows.

ECT uses one temporal support and two condition--time branches, whereas ChordEdit uses two supports and four branches. Both cores make one batched denoiser call and one latent update, so ECT halves the branches within the batch. Refinement adds one sequential noise-and-denoise call to each editor: both use one call without refinement and two with refinement. The smaller core batch reduces denoiser and activation costs at matched budgets (Appendix~\ref{app:resources}, Table~\ref{tab:budget}).

\subsection{User Study}
\label{sec:user_study}

\begin{figure}[tbp]
  \centering
  \includegraphics[width=\linewidth]{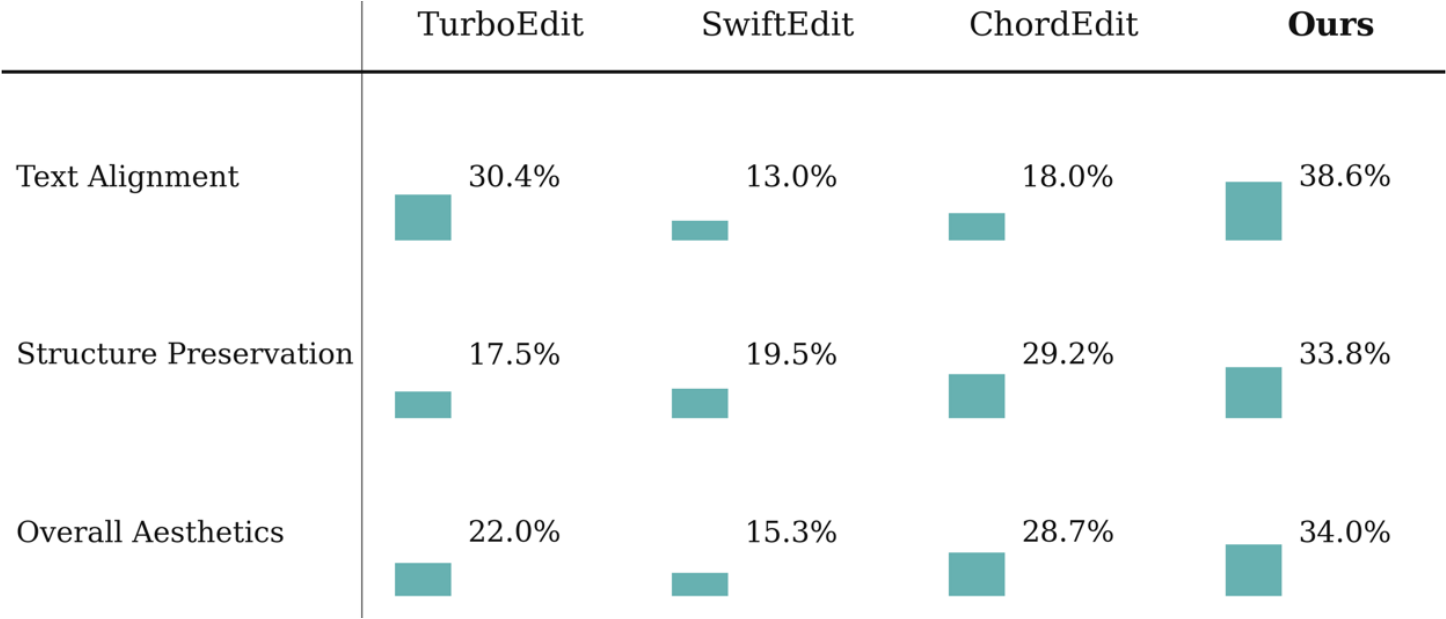}
  \caption{User-study first-place shares (\%), based on mean participant scores per case, with no top ties. ECT leads in text alignment ($38.6\%$), structure preservation ($33.8\%$), and aesthetics ($34.0\%$).}
  \label{fig:user_study}
\end{figure}

We compare TurboEdit, SwiftEdit, ChordEdit, and ECT in a blinded PIE-Bench study. Participants independently score anonymous outputs on assigned subsets of varying sizes from 1 to 5 for text alignment, structure preservation, and aesthetics. Methods are ranked by mean participant scores per case and criterion (Figure~\ref{fig:user_study}). ECT leads with first-place shares of $38.6\%$, $33.8\%$, and $34.0\%$, exceeding runners-up---TurboEdit for alignment and ChordEdit for structure and aesthetics---by $8.2$, $4.6$, and $5.3$ percentage points, respectively. These descriptive differences alone do not establish statistical significance. Appendix~\ref{app:user_study} details the protocol.

\subsection{Ablation Studies}
\label{sec:ablation}
\begin{table*}[tp]
\caption{Ablations on PIE-Bench: step count (a), angular transport (b), and extra NFE (c). Blue shading identifies ECT variants. Panels (a,b) include refinement; the step count in (a) refers to the editing core. Panel (c) compares refinement choices.}
\label{tab:ablations}
\centering
\fontsize{8.5}{10.5}\selectfont
\setlength{\tabcolsep}{3pt}
\renewcommand{\arraystretch}{1.12}
{\raggedright\textbf{(a) One- vs.\ two-step ECT}\par}\vspace{2pt}
\begin{tabularx}{\textwidth}{@{}lXrrrrrr@{}}

                \toprule
                Model & Setting
                & Time (s) $\downarrow$
                & CPU (\%) $\downarrow$
                & Mem. (MiB) $\downarrow$
                & CP $\uparrow$
                & VQ $\uparrow$
                & PC $\uparrow$ \\
                \midrule
                SB-v2 & ChordEdit
                & 0.265 & 357 & 1298.74 & 3.91 & 4.37 & 4.05 \\
                \rowcolor{blue!4}
                SB-v2 & ECT (2-step)
                & 0.259 & 446 & 1382.99 & 3.99 & 4.42 & 4.10 \\
                \rowcolor{blue!8}
                SB-v2 & ECT (1-step)
                & \cellcolor{orange!25}\shortstack[r]{\textbf{0.213}\\[-0.5pt]{\fontsize{6}{7}\selectfont (-17.8\%)}}
                & \cellcolor{orange!25}\shortstack[r]{\textbf{190}\\[-0.5pt]{\fontsize{6}{7}\selectfont (-57.4\%)}}
                & \cellcolor{orange!25}\shortstack[r]{\textbf{1075.93}\\[-0.5pt]{\fontsize{6}{7}\selectfont (-22.2\%)}}
                & \textbf{4.03} & \textbf{4.45} & \textbf{4.13} \\
                \addlinespace
                SD-T & ChordEdit
                & 0.265 & 357 & 1297.07 & 3.95 & 4.37 & 4.00 \\
                \rowcolor{blue!4}
                SD-T & ECT (2-step)
                & 0.259 & 452 & 1386.67 & 4.04 & 4.44 & 4.07 \\
                \rowcolor{blue!8}
                SD-T & ECT (1-step)
                & \cellcolor{orange!25}\shortstack[r]{\textbf{0.212}\\[-0.5pt]{\fontsize{6}{7}\selectfont (-18.1\%)}}
                & \cellcolor{orange!25}\shortstack[r]{\textbf{190}\\[-0.5pt]{\fontsize{6}{7}\selectfont (-57.9\%)}}
                & \cellcolor{orange!25}\shortstack[r]{\textbf{1075.96}\\[-0.5pt]{\fontsize{6}{7}\selectfont (-22.4\%)}}
                & \textbf{4.09} & \textbf{4.65} & \textbf{4.31} \\
                \bottomrule
            
\end{tabularx}\par
\vspace{4mm}
{\raggedright\textbf{(b) Transport stage}\par}\vspace{2pt}
\begin{tabularx}{\textwidth}{@{}lXrrrrr@{}}

                \toprule
                Model & Variant
                & PSNR $\uparrow$
                & LPIPS $\downarrow$
                & CLIP $\uparrow$
                & VQ $\uparrow$
                & PC $\uparrow$ \\
                \midrule
                SB-v2 & w/o transport
                & 22.7 & 124.2 & \textbf{29.8} & 4.4 & 4.0 \\
                \rowcolor{blue!8}
                SB-v2 & ECT (full)
                & \textbf{22.9} & \textbf{122.8} & 29.7
                & \textbf{4.5} & \textbf{4.1} \\
                SD-T & w/o transport
                & 22.4 & 127.1 & \textbf{30.7} & 4.4 & 4.0 \\
                \rowcolor{blue!8}
                SD-T & ECT (full)
                & \textbf{23.7} & \textbf{109.0} & 30.2
                & \textbf{4.7} & \textbf{4.3} \\
                \bottomrule
            
\end{tabularx}\par
\vspace{4mm}
{\raggedright\textbf{(c) Extra NFE}\par}\vspace{2pt}
\begin{tabularx}{\textwidth}{@{}lXrrrrr@{}}

                \toprule
                Model & Variant
                & PSNR $\uparrow$
                & LPIPS $\downarrow$
                & CLIP $\uparrow$
                & VQ $\uparrow$
                & PC $\uparrow$ \\
                \midrule
                SB-v2 & w/o extra NFE
                & \textbf{22.9} & 142.0 & 22.8
                & \textbf{4.5} & \textbf{4.2} \\
                \rowcolor{blue!8}
                SB-v2 & w/ extra NFE
                & \textbf{22.9} & \textbf{122.8} & \textbf{29.7}
                & \textbf{4.5} & 4.1 \\
                SD-T & w/o extra NFE
                & \textbf{26.0} & \textbf{64.1} & 28.6
                & 4.3 & 3.9 \\
                \rowcolor{blue!8}
                SD-T & w/ extra NFE
                & 23.7 & 109.0 & \textbf{30.2}
                & \textbf{4.7} & \textbf{4.3} \\
                \bottomrule
            
\end{tabularx}\par
\end{table*}
\textbf{Step count, transport and refinement.} Table~\ref{tab:ablations} isolates these three components. In panel (a), the one-update core reduces latency by $17.8\%/18.1\%$, CPU utilization by $57.4\%/57.9\%$, and RSS by $22.2\%/22.4\%$ relative to the two-update core on SB-v2/SD-T; CP, VQ and PC also increase. With refinement held fixed, panel (b) improves PSNR and LPIPS on both backbones and raises VQ/PC. Panel (c) shows that refinement reduces SB-v2 LPIPS from $142.0$ to $122.8$ and increases CLIP from $22.8$ to $29.7$; SD-T increases CLIP/VQ/PC with more background change. Here CLIP is the full-image/target-text embedding cosine similarity from ViT-B/32, multiplied by 100; Appendix~\ref{app:image_metrics} defines its computation. These experiments separate the effects of core update count, angular transport and the additional refinement call on both backbones.

\begin{table}[!tbp]

\caption{SD-Turbo scalar ablation at $t=0.75$ without refinement. Raw uses the unscaled endpoint residual. All rules share 700 cases, including 556 valid backgrounds for preservation and BPB.}
\label{tab:scalar_main}
\centering\small
\setlength{\tabcolsep}{2pt}
\begin{tabular}{lrrrr}
\toprule
Update rule & Coefficient & PSNR $\uparrow$ & LPIPS$\times1000$ $\downarrow$ & BPB $\uparrow$ \\
\midrule
Raw & 1.000000 & 23.130 & 100.190 & 24.2550 \\
Fixed 0.7 & 0.700000 & 25.471 & 69.425 & 25.3071 \\
Transport & 1.369884 & 20.995 & 135.810 & 22.8632 \\
ECT & 0.669884 & 25.739 & 66.428 & 25.4060 \\
\bottomrule
\end{tabular}
\end{table}

\textbf{Transport and endpoint calibration.} Table~\ref{tab:scalar_main} reports an additional 700-case run at $t=0.75$, keeping predictions fixed across four update rules. Relative to the raw residual, ECT raises PSNR from $23.130$ to $25.739$, reduces LPIPS from $100.190$ to $66.428$, and raises BPB from $24.2550$ to $25.4060$. Relative to fixed $0.7$, ECT improves PSNR by $0.268$ dB and BPB by $0.0989$. Transport alone enlarges the update; endpoint calibration yields the stronger preservation and BPB values in this comparison. Appendix~\ref{app:cross_time_reproduction} gives all five evaluation times and the encoder check.

The seed control evaluates eleven independent noisy states with refinement enabled. Even the worst seed exceeds matched ChordEdit on all four preservation metrics (Appendix~\ref{app:seed_stability}). The edit-strength sweep gives the best preservation at $\lambda=1$; $\lambda=2$ adds $0.50$ CLIP points with a $2.690$-dB PSNR reduction and $37.636$ higher scaled LPIPS. Appendix~\ref{app:extra_ablations} reports the complete seed, refinement and strength experiments; Table~\ref{tab:ablations} gives the complete transport and step-count ablations; Appendix~\ref{app:legacy_ablations} retains the parameter curves and qualitative transport examples.

\section{Conclusion}
\label{sec:conclusion}

We presented Endpoint-Calibrated Transport (ECT), a framework addressing whether reliable inversion-free editing can be constructed from one temporal field evaluation. Its key is shared-state information in variance-preserving (VP) one-step models: one source--target conditional query determines a local angular semantic tangent and a clean endpoint displacement along the same edit axis. ECT constructs nominal angular transport over the known remaining VP angle and calibrates it through a closed-form endpoint-constrained projection, yielding a transport-then-correction mechanism. The quadrature bound and shrinkage-risk analysis under the stated assumptions provide its theoretical basis. ECT applies directly to pretrained epsilon-prediction VP one-step generators. Its editing core uses one noise level, two text-conditioned branches packed into one batched denoiser call, and one latent update, without additional training, image inversion, a second temporal support, trajectory integration, or iterative optimization. Shared-state information thus supports both semantic editing and structural control while reducing the cost of additional temporal observations.

Experiments across multiple backbones and real-image editing benchmarks show improvements in edited-image quality and the joint evaluation of text alignment and background preservation, alongside lower inference latency, CPU utilization, and host-memory use. These results validate ECT and the practical quality--efficiency value of editing from one temporal support.

\textbf{Limitations and future work.} Current experiments cover four backbones; validation on other backbones remains to be completed. Efficiency measurements cover a limited range of hardware models, and reported values depend on hardware configuration, process scope, and measurement protocol. Future work will extend backbone and hardware coverage and include broader controls.

\subsection*{AI Use Statement}
Generative AI tools were used to assist with language editing, manuscript organization, and consistency checking. The authors reviewed all derivations, implementation correspondences, experimental values, citations, and claims, and take responsibility for the content.

\subsection*{Ethics Statement}
This work includes an anonymous perceptual study of generated images. The rating interface requests no names, email addresses, or demographic attributes; the study database stores a pseudonymous browser identifier, ratings, and UTC timestamps. Like other generative editing systems, the method may be misused to create misleading visual content. Deployment should include provenance, consent, and disclosure practices appropriate to the application.

\subsection*{Reproducibility Statement}
Section~\ref{sec:method} specifies the complete closed-form method and assumptions. Appendix~\ref{app:angular_proof} proves the single-evaluation approximation bound and Appendix~\ref{app:kkt} derives the endpoint projection. Appendices~\ref{app:image_metrics}--\ref{app:user_study} define automatic image metrics, system-resource accounting, ImgEdit-Judge evaluation, and the user-study protocol. Appendix~\ref{app:extra_ablations} reports the seed and $\lambda$ controls together with the exact values used to produce every table. The supplementary material will include configuration files, benchmark preprocessing, per-sample outputs, and commands for both backbones.

\bibliography{iclr2027_conference}

\begin{thebibliography}{43}
\providecommand{\natexlab}[1]{#1}
\providecommand{\url}[1]{\texttt{#1}}
\expandafter\ifx\csname urlstyle\endcsname\relax
  \providecommand{\doi}[1]{doi: #1}\else
  \providecommand{\doi}{doi: \begingroup \urlstyle{rm}\Url}\fi

\bibitem[Beaudouin et~al.(2026)Beaudouin, Li, Kim, Yoon, and
  Wang]{beaudouin2025drfs}
Gaspard Beaudouin, Minghan Li, Jaeyeon Kim, Sung-Hoon Yoon, and Mengyu Wang.
\newblock Delta rectified flow sampling for text-to-image editing.
\newblock In \emph{Proceedings of the IEEE/CVF Conference on Computer Vision
  and Pattern Recognition}, pp.\  18662--18672, 2026.
\newblock URL
  \url{https://github.com/Harvard-AI-and-Robotics-Lab/DeltaRectifiedFlowSampling}.

\bibitem[{Black Forest Labs}(2024)]{bfl2024fluxschnell}
{Black Forest Labs}.
\newblock {FLUX.1 [schnell]}.
\newblock Official model card, 2024.
\newblock URL \url{https://huggingface.co/black-forest-labs/FLUX.1-schnell}.

\bibitem[Brooks et~al.(2023)Brooks, Holynski, and
  Efros]{brooks2023instructpix2pix}
Tim Brooks, Aleksander Holynski, and Alexei~A. Efros.
\newblock Instructpix2pix: Learning to follow image editing instructions, 2023.
\newblock URL \url{https://arxiv.org/abs/2211.09800}.

\bibitem[Cao et~al.(2023)Cao, Wang, Qi, Shan, Qie, and Zheng]{cao2023masactrl}
Mingdeng Cao, Xintao Wang, Zhongang Qi, Ying Shan, Xiaohu Qie, and Yinqiang
  Zheng.
\newblock Masactrl: Tuning-free mutual self-attention control for consistent
  image synthesis and editing.
\newblock In \emph{2023 IEEE/CVF International Conference on Computer Vision
  (ICCV)}, pp.\  22503--22513. IEEE, 2023.

\bibitem[Dao et~al.(2024)Dao, Nguyen, Le, Vu, Nguyen, Pham, and
  Tran]{dao2024swiftbrushv2}
Trung Dao, Thuan~Hoang Nguyen, Thanh Le, Duc Vu, Khoi Nguyen, Cuong Pham, and
  Anh Tran.
\newblock Swiftbrush v2: Make your one-step diffusion model better than its
  teacher.
\newblock In \emph{European Conference on Computer Vision}, pp.\  176--192.
  Springer, 2024.

\bibitem[Deutch et~al.(2024)Deutch, Gal, Garibi, Patashnik, and
  Cohen-Or]{deutch2024turboedit}
Gilad Deutch, Rinon Gal, Daniel Garibi, Or~Patashnik, and Daniel Cohen-Or.
\newblock Turboedit: Text-based image editing using few-step diffusion models.
\newblock In \emph{SIGGRAPH Asia 2024 Conference Papers}, pp.\  1--12, 2024.

\bibitem[Esser et~al.(2024)Esser, Kulal, Blattmann, Entezari, M{\"u}ller,
  Saini, Levi, Lorenz, Sauer, Boesel, et~al.]{esser2024sd3}
Patrick Esser, Sumith Kulal, Andreas Blattmann, Rahim Entezari, Jonas
  M{\"u}ller, Harry Saini, Yam Levi, Dominik Lorenz, Axel Sauer, Frederic
  Boesel, et~al.
\newblock Scaling rectified flow transformers for high-resolution image
  synthesis.
\newblock In \emph{Forty-first international conference on machine learning},
  2024.

\bibitem[Gong et~al.(2025)Gong, Zhu, and Zhang]{gong2025instantedit}
Yiming Gong, Zhen Zhu, and Minjia Zhang.
\newblock Instantedit: Text-guided few-step image editing with piecewise
  rectified flow.
\newblock In \emph{2025 IEEE/CVF International Conference on Computer Vision
  (ICCV)}, pp.\  16808--16817. IEEE, 2025.

\bibitem[Greenberg(2025)]{greenberg2025flux}
Or~Greenberg.
\newblock Demystifying flux architecture.
\newblock \emph{arXiv preprint arXiv:2507.09595}, 2025.

\bibitem[Hertz et~al.(2022)Hertz, Mokady, Tenenbaum, Aberman, Pritch, and
  Cohen-Or]{hertz2023prompt}
Amir Hertz, Ron Mokady, Jay Tenenbaum, Kfir Aberman, Yael Pritch, and Daniel
  Cohen-Or.
\newblock Prompt-to-prompt image editing with cross attention control.
\newblock \emph{arXiv preprint arXiv:2208.01626}, 2022.

\bibitem[Ho \& Salimans(2022)Ho and Salimans]{ho2022cfg}
Jonathan Ho and Tim Salimans.
\newblock Classifier-free diffusion guidance.
\newblock \emph{arXiv preprint arXiv:2207.12598}, 2022.

\bibitem[Ho et~al.(2020)Ho, Jain, and Abbeel]{ho2020ddpm}
Jonathan Ho, Ajay Jain, and Pieter Abbeel.
\newblock Denoising diffusion probabilistic models.
\newblock \emph{Advances in neural information processing systems},
  33:\penalty0 6840--6851, 2020.

\bibitem[Jiang et~al.(2026)Jiang, Wu, Pei, Wu, Chen, Zhao, Yu, and
  Cai]{jiang2026snredit}
Lifan Jiang, Boxi Wu, Yuhang Pei, Tianrun Wu, Yongyuan Chen, Yan Zhao, Shiyu
  Yu, and Deng Cai.
\newblock Snr-edit: Structure-aware noise rectification for inversion-free
  flow-based editing.
\newblock \emph{arXiv preprint arXiv:2601.19180}, 2026.

\bibitem[Kim et~al.(2026)Kim, Hong, Park, and Ye]{kim2026flowalign}
Jeongsol Kim, Yeobin Hong, Jonghyun Park, and Jong~Chul Ye.
\newblock Flowalign: Trajectory-regularized, inversion-free flow-based image
  editing.
\newblock In \emph{International Conference on Learning Representations},
  volume 2026, pp.\  121814--121846, 2026.

\bibitem[Kulikov et~al.(2025)Kulikov, Kleiner, Huberman-Spiegelglas, and
  Michaeli]{kulikov2025flowedit}
Vladimir Kulikov, Matan Kleiner, Inbar Huberman-Spiegelglas, and Tomer
  Michaeli.
\newblock Flowedit: Inversion-free text-based editing using pre-trained flow
  models.
\newblock In \emph{2025 IEEE/CVF International Conference on Computer Vision
  (ICCV)}, pp.\  19721--19730. IEEE, 2025.

\bibitem[Li et~al.(2026)Li, Moebel, and Wang]{li2026rethinking}
Minghan Li, Jeremy Moebel, and Mengyu Wang.
\newblock Rethinking one-step image editing through chordedit: Reproduction,
  simplification, and new insights.
\newblock \emph{arXiv preprint arXiv:2606.14042}, 2026.

\bibitem[Lipman et~al.(2022)Lipman, Chen, Ben-Hamu, Nickel, and
  Le]{lipman2023flowmatching}
Yaron Lipman, Ricky~TQ Chen, Heli Ben-Hamu, Maximilian Nickel, and Matt Le.
\newblock Flow matching for generative modeling.
\newblock \emph{arXiv preprint arXiv:2210.02747}, 2022.

\bibitem[Liu et~al.(2022)Liu, Gong, and Liu]{liu2023rectifiedflow}
Xingchao Liu, Chengyue Gong, and Qiang Liu.
\newblock Flow straight and fast: Learning to generate and transfer data with
  rectified flow, 2022.
\newblock URL \url{https://arxiv.org/abs/2209.03003}.

\bibitem[Liu et~al.(2024)Liu, Zhang, Ma, Peng, and Liu]{liu2024instaflow}
Xingchao Liu, Xiwen Zhang, Jianzhu Ma, Jian Peng, and Qiang Liu.
\newblock Instaflow: One step is enough for high-quality diffusion-based
  text-to-image generation.
\newblock In B.~Kim, Y.~Yue, S.~Chaudhuri, K.~Fragkiadaki, M.~Khan, and Y.~Sun
  (eds.), \emph{International Conference on Learning Representations}, volume
  2024, pp.\  17860--17889, 2024.
\newblock URL
  \url{https://proceedings.iclr.cc/paper_files/paper/2024/file/4dc37a7bc61057252ce043fa3b83aac2-Paper-Conference.pdf}.

\bibitem[Lu et~al.(2026)Lu, Chen, Guo, Li, Wang, and Shi]{lu2026chordedit}
Liangsi Lu, Xuhang Chen, Minzhe Guo, Shichu Li, Jingchao Wang, and Yang Shi.
\newblock Chordedit: One-step low-energy transport for image editing.
\newblock \emph{arXiv preprint arXiv:2602.19083}, 2026.

\bibitem[Lüddecke \& Ecker(2022)Lüddecke and Ecker]{luddecke2022clipseg}
Timo Lüddecke and Alexander Ecker.
\newblock Image segmentation using text and image prompts.
\newblock In \emph{Proceedings of the IEEE/CVF Conference on Computer Vision
  and Pattern Recognition}, pp.\  7086--7096, 2022.

\bibitem[Meng et~al.(2022)Meng, He, Song, Song, Wu, Zhu, and
  Ermon]{meng2022sdedit}
Chenlin Meng, Yutong He, Yang Song, Jiaming Song, Jiajun Wu, Jun-Yan Zhu, and
  Stefano Ermon.
\newblock Sdedit: guided image synthesis and editing with stochastic
  differential equations (2022).
\newblock \emph{arXiv preprint arXiv:2108.01073}, 2022.

\bibitem[Miyake et~al.(2025)Miyake, Iohara, Saito, and
  Tanaka]{miyake2025negative}
Daiki Miyake, Akihiro Iohara, Yu~Saito, and Toshiyuki Tanaka.
\newblock Negative-prompt inversion: Fast image inversion for editing with
  text-guided diffusion models.
\newblock In \emph{2025 IEEE/CVF Winter Conference on Applications of Computer
  Vision (WACV)}, pp.\  2063--2072. IEEE, 2025.

\bibitem[Mokady et~al.(2023)Mokady, Hertz, Aberman, Pritch, and
  Cohen-Or]{mokady2023null}
Ron Mokady, Amir Hertz, Kfir Aberman, Yael Pritch, and Daniel Cohen-Or.
\newblock Null-text inversion for editing real images using guided diffusion
  models.
\newblock In \emph{2023 IEEE/CVF Conference on Computer Vision and Pattern
  Recognition (CVPR)}, pp.\  6038--6047. IEEE, 2023.

\bibitem[Nguyen \& Tran(2024)Nguyen and Tran]{nguyen2024swiftbrush}
Thuan~Hoang Nguyen and Anh Tran.
\newblock Swiftbrush: One-step text-to-image diffusion model with variational
  score distillation.
\newblock In \emph{2024 IEEE/CVF Conference on Computer Vision and Pattern
  Recognition (CVPR)}, pp.\  7807--7816. IEEE, 2024.

\bibitem[Nguyen et~al.(2025)Nguyen, Nguyen, Nguyen, Tran, and
  Pham]{nguyen2025swiftedit}
Trong-Tung Nguyen, Quang Nguyen, Khoi Nguyen, Anh Tran, and Cuong Pham.
\newblock Swiftedit: Lightning fast text-guided image editing via one-step
  diffusion.
\newblock In \emph{2025 IEEE/CVF Conference on Computer Vision and Pattern
  Recognition (CVPR)}, pp.\  21492--21501. IEEE, 2025.

\bibitem[Podell et~al.(2023)Podell, English, Lacey, Blattmann, Dockhorn,
  Müller, Penna, and Rombach]{podell2024sdxl}
Dustin Podell, Zion English, Kyle Lacey, Andreas Blattmann, Tim Dockhorn, Jonas
  Müller, Joe Penna, and Robin Rombach.
\newblock Sdxl: Improving latent diffusion models for high-resolution image
  synthesis, 2023.
\newblock URL \url{https://arxiv.org/abs/2307.01952}.

\bibitem[Radford et~al.(2021)Radford, Kim, Hallacy, Ramesh, Goh, Agarwal,
  Sastry, Askell, Mishkin, Clark, et~al.]{radford2021clip}
Alec Radford, Jong~Wook Kim, Chris Hallacy, Aditya Ramesh, Gabriel Goh,
  Sandhini Agarwal, Girish Sastry, Amanda Askell, Pamela Mishkin, Jack Clark,
  et~al.
\newblock Learning transferable visual models from natural language
  supervision.
\newblock In \emph{International conference on machine learning}, pp.\
  8748--8763. PmLR, 2021.

\bibitem[Rombach et~al.(2022)Rombach, Blattmann, Lorenz, Esser, and
  Ommer]{rombach2022ldm}
Robin Rombach, Andreas Blattmann, Dominik Lorenz, Patrick Esser, and Bj{\"o}rn
  Ommer.
\newblock High-resolution image synthesis with latent diffusion models.
\newblock In \emph{2022 IEEE/CVF conference on computer vision and pattern
  recognition (CVPR)}, pp.\  10674--10685. ieee, 2022.

\bibitem[Sauer et~al.(2024)Sauer, Lorenz, Blattmann, and Rombach]{sauer2024add}
Axel Sauer, Dominik Lorenz, Andreas Blattmann, and Robin Rombach.
\newblock Adversarial diffusion distillation.
\newblock In \emph{European Conference on Computer Vision}, pp.\  87--103.
  Springer, 2024.

\bibitem[Song et~al.(2020)Song, Meng, and Ermon]{song2021ddim}
Jiaming Song, Chenlin Meng, and Stefano Ermon.
\newblock Denoising diffusion implicit models.
\newblock \emph{arXiv preprint arXiv:2010.02502}, 2020.

\bibitem[Song et~al.(2023)Song, Dhariwal, Chen, and
  Sutskever]{song2023consistency}
Yang Song, Prafulla Dhariwal, Mark Chen, and Ilya Sutskever.
\newblock Consistency models.
\newblock \emph{arXiv preprint arXiv:2303.01469}, 2023.

\bibitem[{Stability AI}(2024)]{stabilityai2024sd35}
{Stability AI}.
\newblock {Stable Diffusion 3.5 Large}.
\newblock Official model card, 2024.
\newblock URL
  \url{https://huggingface.co/stabilityai/stable-diffusion-3.5-large}.

\bibitem[Tumanyan et~al.(2023)Tumanyan, Geyer, Bagon, and
  Dekel]{tumanyan2023plug}
Narek Tumanyan, Michal Geyer, Shai Bagon, and Tali Dekel.
\newblock Plug-and-play diffusion features for text-driven image-to-image
  translation.
\newblock In \emph{2023 IEEE/CVF Conference on Computer Vision and Pattern
  Recognition (CVPR)}, pp.\  1921--1930. IEEE, 2023.

\bibitem[Wang et~al.(2004)Wang, Bovik, Sheikh, and Simoncelli]{wang2004ssim}
Zhou Wang, Alan~C Bovik, Hamid~R Sheikh, and Eero~P Simoncelli.
\newblock Image quality assessment: from error visibility to structural
  similarity.
\newblock \emph{IEEE transactions on image processing}, 13\penalty0
  (4):\penalty0 600--612, 2004.

\bibitem[Xie et~al.(2025)Xie, Li, Li, Wu, Yi, and Zhang]{xie2025dnaedit}
Chenxi Xie, Minghan Li, Shuai Li, Yuhui Wu, Qiaosi Yi, and Lei Zhang.
\newblock Dnaedit: Direct noise alignment for text-guided rectified flow
  editing.
\newblock In D.~Belgrave, C.~Zhang, H.~Lin, R.~Pascanu, P.~Koniusz,
  M.~Ghassemi, and N.~Chen (eds.), \emph{Advances in Neural Information
  Processing Systems}, volume 38, Main Conference, pp.\  124456--124474. Curran
  Associates, Inc., 2025.
\newblock \doi{10.52202/085713-4147}.
\newblock URL
  \url{https://proceedings.neurips.cc/paper_files/paper/2025/file/b44ae90136013a8d0e2d24f6015b6097-Paper-Conference.pdf}.

\bibitem[Xu et~al.(2025)Xu, Jiang, Hu, Luo, He, Zhang, Wang, Wu, Ling, and
  Wang]{xu2025ftedit}
Pengcheng Xu, Boyuan Jiang, Xiaobin Hu, Donghao Luo, Qingdong He, Jiangning
  Zhang, Chengjie Wang, Yunsheng Wu, Charles Ling, and Boyu Wang.
\newblock Unveil inversion and invariance in flow transformer for versatile
  image editing.
\newblock In \emph{2025 IEEE/CVF Conference on Computer Vision and Pattern
  Recognition (CVPR)}, pp.\  28479--28489. IEEE, 2025.

\bibitem[Xu et~al.(2024)Xu, Huang, Pan, Ma, and Chai]{xu2024infedit}
Sihan Xu, Yidong Huang, Jiayi Pan, Ziqiao Ma, and Joyce Chai.
\newblock Inversion-free image editing with language-guided diffusion models.
\newblock In \emph{2024 IEEE/CVF Conference on Computer Vision and Pattern
  Recognition (CVPR)}, pp.\  9454--9461. IEEE, 2024.

\bibitem[Yan et~al.(2024)Yan, Liu, Pan, Liew, Liu, and Feng]{yan2024perflow}
Hanshu Yan, Xingchao Liu, Jiachun Pan, Jun~Hao Liew, Qiang Liu, and Jiashi
  Feng.
\newblock Perflow: Piecewise rectified flow as universal plug-and-play
  accelerator.
\newblock \emph{Advances in Neural Information Processing Systems},
  37:\penalty0 78630--78652, 2024.

\bibitem[Yang \& Ye(2026)Yang and Ye]{yang2026directedit}
Desong Yang and Mang Ye.
\newblock Directedit: Step-level accurate inversion for flow-based image
  editing.
\newblock \emph{arXiv preprint arXiv:2605.02417}, 2026.

\bibitem[Yang et~al.(2026)Yang, Shen, Li, Dai, Luo, Ma, Fang, Li, and
  Wang]{yang2026fiaedit}
Kaixiang Yang, Boyang Shen, Xin Li, Yuchen Dai, Yuxuan Luo, Yueran Ma, Wei
  Fang, Qiang Li, and Zhiwei Wang.
\newblock Fia-edit: Frequency-interactive attention for efficient and
  high-fidelity inversion-free text-guided image editing.
\newblock In \emph{Proceedings of the AAAI Conference on Artificial
  Intelligence}, volume~40, pp.\  11613--11621, 2026.
\newblock No. 14.

\bibitem[Ye et~al.(2025)Ye, He, Li, Lin, Yuan, Yan, Hou, and
  Yuan]{ye2025imgedit}
Yang Ye, Xianyi He, Zongjian Li, Bin Lin, Shenghai Yuan, Zhiyuan Yan, Bohan
  Hou, and Li~Yuan.
\newblock Imgedit: A unified image editing dataset and benchmark.
\newblock In D.~Belgrave, C.~Zhang, H.~Lin, R.~Pascanu, P.~Koniusz,
  M.~Ghassemi, and N.~Chen (eds.), \emph{Advances in Neural Information
  Processing Systems}, volume 38, Main Conference. Curran Associates, Inc.,
  2025.
\newblock \doi{10.52202/085713-4387}.
\newblock URL
  \url{https://proceedings.neurips.cc/paper_files/paper/2025/file/bec783f119b969772c35243b1c801090-Paper-Datasets_and_Benchmarks_Track.pdf}.

\bibitem[Zhang et~al.(2018)Zhang, Isola, Efros, Shechtman, and
  Wang]{zhang2018lpips}
Richard Zhang, Phillip Isola, Alexei~A Efros, Eli Shechtman, and Oliver Wang.
\newblock The unreasonable effectiveness of deep features as a perceptual
  metric.
\newblock In \emph{2018 IEEE/CVF conference on computer vision and pattern
  recognition}, pp.\  586--595. IEEE, 2018.

\end{thebibliography}
\bibliographystyle{iclr2027_conference}

\onecolumn
\appendix

\section{Angular Derivations and Approximation Error}
\label{app:angular_proof}

Differentiating the VP angular path in Equation~\ref{eq:angular_parameterization} gives
\begin{equation}
    \frac{dz_\theta}{d\theta}=\cos\theta\,\epsilon-\sin\theta\,x_0,
\end{equation}
which yields Equation~\ref{eq:endpoint_velocity} after substituting the model estimates. For the quadrature bound, write
\begin{equation}
    a_t^*=\int_{\theta_t}^{0}g(\theta)d\theta,
    \qquad
    \widehat a_t=-\theta_tg(\theta_t).
\end{equation}
If $\|g(\theta_1)-g(\theta_2)\|_2\le L|\theta_1-\theta_2|$, then
\begin{align}
    \|a_t^*-\widehat a_t\|_2
    &\le \int_0^{\theta_t}\|g(\theta)-g(\theta_t)\|_2d\theta \\
    &\le \int_0^{\theta_t}L(\theta_t-\theta)d\theta
    =\frac{L}{2}\theta_t^2.
\end{align}

\paragraph{Angular-horizon interpretation.}
The nominal transport rescales the naive one-step edit by a factor fixed entirely by the schedule. From the shared state (Equation~\ref{eq:shared_state_residuals}), the single-step endpoint displacement has magnitude $\propto\sigma_t/\alpha_t$ while the transport has magnitude $\propto\theta_t/\alpha_t$, so
\begin{equation}
    \|\widehat a_t\|_2=\frac{\theta_t}{\sigma_t}\|d_t\|_2
    =\frac{\theta_t}{\sin\theta_t}\|d_t\|_2\ge\|d_t\|_2 .
\end{equation}
Under the frozen-tangent approximation, this factor replaces $\sin\theta_t$ with $\theta_t$ and tends to one as $\theta_t\to0$. It relates the VP angular horizon to the one-step chord under that approximation.

\paragraph{Prediction error and shrinkage.}
For a model residual $g_t$ with $\|g_t-g(\theta_t)\|_2\le\eta$, the triangle inequality gives $\|a_t^*+\theta_tg_t\|_2\le L\theta_t^2/2+\theta_t\eta$. The first term describes angular quadrature error and the second accounts for predictor error. The shrinkage analysis below assumes unbiased noise.

\section{Endpoint Projection and KKT Conditions}
\label{app:kkt}

The Lagrangian of Equation~\ref{eq:ect_projection} is
\begin{equation}
    \mathcal{L}(u,\zeta)
    =\frac{1}{2}\|u-\widehat a_t\|_2^2
    +\zeta(\langle d_t,u\rangle-b_t),
    \qquad \zeta\ge0.
\end{equation}
Stationarity gives $u_t^*=\widehat a_t-\zeta d_t$; for active constraints ($d_t\ne0$), complementary slackness gives
\begin{equation}
    \zeta_t^*
    =\frac{\langle d_t,\widehat a_t\rangle-b_t}{\|d_t\|_2^2},
    \qquad
    b_t=\langle d_t,\widehat a_t\rangle-\zeta_t^*\|d_t\|_2^2.
\end{equation}
For the degenerate case $d_t=0$, we bypass the quotient and set $\zeta=0$. Thus a fixed $\zeta$ is equivalent to a sample-dependent implicit exposure budget. Conversely, a specified budget yields an adaptive dual strength from the same observation.

\section{Shrinkage-Risk Interpretation}
\label{app:risk}

For the final editor in Equation~\ref{eq:closed_form}, write $u=\kappa d_t$ with $\kappa=\lambda(\theta_t/\sigma_t-\zeta)$. Assume $d_t=d^*+e$, with $\mathbb{E}[e]=0$, $\mathbb{E}\|e\|_2^2=V$, and $\|d^*\|_2^2=S$. Then $u$ has risk
\begin{equation}
    \mathbb{E}\|u-d^*\|_2^2=(1-\kappa)^2S+\kappa^2V.
\end{equation}
For $0<\kappa<1$, this is lower than the uncorrected risk $V$ whenever
\begin{equation}
    \frac{V}{S}>\frac{1-\kappa}{1+\kappa}.
\end{equation}
This result compares the final coefficient $\kappa d_t$ with the unscaled estimator $d_t$ for the target $d^*$. Under the stated noise assumptions, it motivates conservative endpoint calibration relative to the unscaled endpoint estimator.

\section{Automatic Image Metrics}
\label{app:image_metrics}

\paragraph{Common preprocessing and scope.}
Let $S,T\in[0,1]^{3\times H\times W}$ denote the source and edited RGB images. If the prediction is larger than the source, we use the benchmark-compatible bottom-right crop; an incompatible size is excluded. Let $M\in\{0,1\}^{H\times W}$ be the edit mask ($M=1$ in the intended edit region). Following the evaluator, all four image boundaries are also set to one before defining $B=1-M$. The preservation inputs are $S_B=S\odot B$ and $T_B=T\odot B$, with $B$ broadcast over channels. MSE, PSNR, SSIM, and LPIPS measure \emph{background preservation}; CLIP measures \emph{target alignment} on the full edited image; CLIP-BPB additionally weights it by protected-content retention (Equation~\ref{eq:bpb}). Samples with no evaluable background are invalid.

\paragraph{Definitions.}
The evaluator computes
\begin{align}
    \operatorname{MSE}(S_B,T_B)
    &=\frac{1}{3HW}\sum_{c,h,w}(S_{B,c,h,w}-T_{B,c,h,w})^2, \\
    \operatorname{PSNR}(S_B,T_B)
    &=10\log_{10}\frac{1}{\operatorname{MSE}(S_B,T_B)}.
\end{align}
The denominator is the full $3HW$, not the number of background elements: the edit region is zero in both inputs and contributes no error but remains in the normalization. PSNR is in dB because the data range is one; an exact match has MSE $0$ and theoretical PSNR $+\infty$.

For corresponding local windows $x$ and $y$, SSIM is~\citep{wang2004ssim}
\begin{equation}
    \operatorname{SSIM}(x,y)=
    \frac{(2\mu_x\mu_y+C_1)(2\sigma_{xy}+C_2)}
    {(\mu_x^2+\mu_y^2+C_1)(\sigma_x^2+\sigma_y^2+C_2)},
\end{equation}
with $C_j=(K_jL)^2$ and $L=1$; the implementation averages over windows and channels. Windows crossing the mask boundary can therefore reflect boundary structure. For LPIPS, define $\widetilde S_B=2S_B-1$ and $\widetilde T_B=2T_B-1$. In its standard notation~\citep{zhang2018lpips},
\begin{equation}
    \operatorname{LPIPS}(\widetilde S_B,\widetilde T_B)=
    \sum_l\frac{1}{H_lW_l}\sum_{h,w}
    \left\lVert w_l\odot
    \left(\hat\phi_l(\widetilde S_B)_{h,w}
    -\hat\phi_l(\widetilde T_B)_{h,w}\right)
    \right\rVert_2^2,
\end{equation}
where $\hat\phi_l$ is the channel-normalized feature at layer $l$ and $w_l$ is its learned channel weight; our evaluator uses the SqueezeNet backbone and released learned weights. CLIP uses cosine similarity, displayed on a $\times100$ scale in the tables:
\begin{equation}
    \operatorname{CLIP}(T,e)=
    \frac{v_I(T)^\top v_T(e)}{\lVert v_I(T)\rVert_2\lVert v_T(e)\rVert_2},
\end{equation}
Here the embeddings represent the full edited image and target instruction $e$, computed with \texttt{openai/\allowbreak clip-\allowbreak vit-\allowbreak base-\allowbreak patch32}~\citep{radford2021clip}. We select $e$ from the first non-empty field in this order: \texttt{editing\_prompt}, \texttt{edited\_prompt}, \texttt{target\_prompt}, then \texttt{editing\_instruction} if the preceding fields are empty.

\begin{table}[htbp]
\caption{Automatic image metrics and their interpretation.}
    \label{tab:metric_summary}
    \centering
    \small
    \setlength{\tabcolsep}{4pt}
    \begin{tabular}{lcll}
        \toprule
        Metric & Direction & Input & Primary interpretation \\
        \midrule
        MSE & $\downarrow$ & Masked source/output & Mean pixel error \\
        PSNR & $\uparrow$ & Masked source/output & Log pixel fidelity \\
        SSIM & $\uparrow$ & Masked source/output & Local structure preservation \\
        LPIPS & $\downarrow$ & Masked source/output & Learned perceptual distance \\
        CLIP & $\uparrow$ & Full output/instruction & Target semantic alignment \\
CLIP-BPB & $\uparrow$ & Source/output/instruction & Joint semantics/preservation \\
        \bottomrule
    \end{tabular}
\end{table}

\paragraph{Aggregation.}
Each metric is computed per sample and then averaged over its finite values, both overall and per edit category. NaN or failed values are excluded rather than replaced by zero; if a metric has no valid samples, its aggregate is NaN. CLIP-BPB combines target-text similarity and background retention; its valid set and coefficient are specified in Appendix~\ref{app:region_protocol}. For display, tables scale MSE and LPIPS by $1000$, and SSIM and CLIP by $100$.

\section{Inference-Time and Resource Accounting}
\label{app:resources}

\begin{table}[!htbp]

\caption{Execution budgets. Supports/branches count the editing core; refinement adds one sequential call. Counts are direct traces, not latency estimates.}
\label{tab:budget}
\centering\small
\begin{tabular}{lrrrr}
\toprule
Configuration & Core supports & Core branches & Core calls & Total calls \\
\midrule
ChordEdit$^0$ & 2 & 4 & 1 & 1 \\
ECT$^0$ & 1 & 2 & 1 & 1 \\
ChordEdit$^+$ & 2 & 4 & 1 & 2 \\
ECT$^+$ & 1 & 2 & 1 & 2 \\
\bottomrule
\end{tabular}
\end{table}

Table~\ref{tab:budget} separates the work performed by each editor. ECT uses one temporal support and two condition--time branches; ChordEdit uses two supports and four branches. Each core executes one batched call and one latent update. With refinement, both execute two sequential calls in total. The smaller core batch explains latency and CPU/RSS savings at matched refinement budgets.

Original resource records use NVIDIA H20; added PIE F-S/SD3.5 and native controls and the fresh scalar reproduction use NVIDIA A800-SXM4-80GB. Hardware groups are marked in Tables~\ref{tab:pie_results}--\ref{tab:snr_results}; paired efficiency comparisons use the same device.

\paragraph{Measurement window.}
For sample $i$, the high-resolution timer surrounds the full editing pipeline call. A CUDA synchronization is issued before recording $t_i^{\mathrm{end}}$, so $d_i=t_i^{\mathrm{end}}-t_i^{\mathrm{start}}$ includes completed GPU work. The window excludes source-image loading, tensor-to-image conversion, output writing, logging, and metric evaluation. Skipped or failed calls are omitted. Mean latency is
\begin{equation}
    \bar d=\frac{1}{N}\sum_{i=1}^{N}d_i\quad\text{seconds per image}.
\end{equation}

\paragraph{CPU and host memory.}
The process is sampled at approximately $0.1$-s intervals during the measured inference windows. After priming \texttt{psutil.Process.cpu\_percent}, the reported CPU value is the time-weighted process utilization
\begin{equation}
    \bar p=\frac{\sum_k p_k\Delta t_k}{\sum_k\Delta t_k}.
\end{equation}
The process convention assigns 100\% to one fully occupied logical core; hence a multithreaded process can exceed 100\%. This is not the 0--100\% system-wide CPU percentage. Mem.~is the process resident-set size (RSS), approximated as a left-endpoint time average,
\begin{equation}
    \bar m=\frac{\sum_k m_{k-1}\Delta t_k}{\sum_k\Delta t_k},
    \qquad \operatorname{MiB}(\bar m)=\bar m/2^{20}.
\end{equation}
This is process-resident memory, not system-wide memory or a percentage of installed RAM.

\paragraph{GPU memory.}
After loading, we record PyTorch's allocated and reserved baselines. Immediately before the first measured inference it resets CUDA peak statistics; after all measured samples it reads \texttt{max\_memory\_allocated} and \texttt{max\_memory\_reserved}. VRAM in the main tables is the latter divided by $2^{20}$, i.e., the peak memory held by the PyTorch caching allocator, including cached blocks. It is therefore generally no smaller than live-tensor allocated memory and is neither mean VRAM occupancy nor GPU compute utilization. We also record the allocated peak and its increase over the model-loaded baseline; VRAM reports only the reserved peak.

\begin{table}[htbp]
\caption{Resource metrics reported in the main tables.}
    \label{tab:resource_summary}
    \centering
    \small
    \setlength{\tabcolsep}{5pt}
    \begin{tabular}{llll}
        \toprule
        Column & Statistic & Scope & Unit \\
        \midrule
        Time & Arithmetic mean of $d_i$ & Pipeline call & s/image \\
        CPU & Time-weighted process utilization & Inference windows & core-\% \\
        Mem. & Time-weighted process RSS & Inference windows & MiB \\
        VRAM & Peak PyTorch reserved memory & Measured run & MiB \\
        \bottomrule
    \end{tabular}
\end{table}

\paragraph{Computational structure and comparability.}
We separately count (i) sequential batched denoiser calls, (ii) distinct temporal supports, (iii) condition--time branches inside the batch, and (iv) latent updates. ECT uses $(1,1,2,1)$ under this ordering, whereas matched ChordEdit uses $(1,2,4,1)$. These are core counts without refinement. Enabling the extra noise-and-denoise stage adds one sequential call to both editors (Table~\ref{tab:budget}); the seed and $\lambda$ scans use refinement. Resource comparisons match hardware, precision, resolution, batch construction, software stack, synchronization, process scope, and sample order.

\section{ImgEdit-Judge Evaluation Protocol}
\label{app:imgedit_judge}

We use the released ImgEdit-Judge checkpoint~\citep{ye2025imgedit}, a Qwen2.5-VL-7B evaluator fine-tuned for image-editing assessment, with one fixed three-dimension scoring prompt for all methods. It is loaded in bfloat16 with automatic device placement; Flash Attention 2 is used when available and SDPA otherwise. Decoding is deterministic (\texttt{do\_sample=False}) with at most 512 new tokens, and the processor uses \texttt{min\_pixels=1,016,064} and \texttt{max\_pixels=1,354,752}.

\paragraph{Input and rubric.}
For sample $i$, the judge receives the scoring prompt with the edit instruction inserted, followed by the original image and edited image:
\begin{equation}
    \mathcal{J}_i=(P(e_i),I_i^{\mathrm{src}},I_i^{\mathrm{edit}}).
\end{equation}
It returns one integer in $\{1,2,3,4,5\}$ for each independent dimension in Table~\ref{tab:judge_rubric}. Requested changes are not penalized under CP, and VQ and PC are scored independently of whether the requested edit was completed. The CLIP term in BPB measures target-text alignment.

\begin{table}[htbp]
\caption{ImgEdit-Judge scoring anchors.}
    \label{tab:judge_rubric}
    \centering
    \footnotesize
    \renewcommand{\arraystretch}{1.08}
    \setlength{\tabcolsep}{6pt}
    \begin{tabular}{@{}>{\raggedright\arraybackslash}p{0.30\linewidth}>{\raggedright\arraybackslash}p{0.30\linewidth}>{\raggedright\arraybackslash}p{0.30\linewidth}@{}}
        \toprule
        \textbf{Score 1: poor} & \textbf{Score 3: adequate} & \textbf{Score 5: excellent} \\
        \midrule
        \rowcolor{blue!8}\multicolumn{3}{c}{\textbf{CP: Content and Identity Preservation}} \\
        Severe unintended loss, deformation, or identity/layout change. & Source remains recognizable, with visible unintended local changes. & Unrequested identity, geometry, layout, and detail are preserved. \\
        \addlinespace[3pt]
        \rowcolor{blue!8}\multicolumn{3}{c}{\textbf{VQ: Visual Quality and Artifact Control}} \\
        Severe blur, noise, corruption, or deformation. & Usable output with visible local blur, noise, or edge artifacts. & Sharp, detailed output without visible editing artifacts. \\
        \addlinespace[3pt]
        \rowcolor{blue!8}\multicolumn{3}{c}{\textbf{PC: Physical and Perceptual Coherence}} \\
        Severe conflicts in lighting, perspective, anatomy, scale, depth, material, or style. & Broadly plausible, with minor physical or perceptual inconsistency. & Edited content is naturally integrated and internally coherent. \\
        \bottomrule
    \end{tabular}
\end{table}

\paragraph{Parsing and aggregation.}
The requested response contains a brief rationale and the three named scores. We strip Markdown markers and parse either the full dimension names or the fallback labels \texttt{Score 1--3}. A missing or out-of-range value is recorded as missing, never as zero. For dimension $q$, with valid set $V_q$, the reported value is
\begin{equation}
    \bar s_q=\frac{1}{|V_q|}\sum_{i\in V_q}s_{i,q},
    \qquad s_{i,q}\in\{1,2,3,4,5\}.
\end{equation}
Thus CP, VQ, and PC may have different denominators. We retain each count and list of missing cases, report the three means separately, and do not form an undocumented composite judge score.

\section{User-Study Protocol}
\label{app:user_study}

\paragraph{Design.}
Participants evaluate assigned subsets of PIE-Bench cases; the number of assigned cases need not be identical across participants. Each case contains the source image, edit instruction, and outputs from TurboEdit, SwiftEdit, ChordEdit, and ECT. The comparison uses cases with all four outputs. Each participant rates the four methods on the three criteria below, and the analysis averages the available participant ratings within each case before comparing methods. Consequently, cases have equal weight in the final win rate even when they receive different numbers of ratings.

\paragraph{Blinding and task.}
The four outputs are presented anonymously, with method identities withheld. Each output receives independent integer scores from 1 (worst) to 5 (best) on the three criteria in Table~\ref{tab:user_study_dimensions}, giving 12 ratings for a completed participant--case record. A participant may assign the same integer score to multiple outputs; the final ranking uses the mean across participants, and no ties for the highest case-level mean occurred in the reported results.

\begin{table}[htbp]
\caption{Human rating dimensions.}
    \label{tab:user_study_dimensions}
    \centering
    \small
    \setlength{\tabcolsep}{5pt}
    \begin{tabular}{p{0.22\linewidth}p{0.68\linewidth}}
        \toprule
        Dimension & Rating criterion \\
        \midrule
        Text alignment & Agreement between the edited image and instruction \\
        Structure preservation & Preservation of non-target content, identity, geometry, and layout \\
        Overall aesthetics & Naturalness, detail, consistency, and artifact control \\
        \bottomrule
    \end{tabular}
\end{table}

\paragraph{Case-level aggregation.}
Let $s_{p,i,m,k}\in\{1,\ldots,5\}$ be participant $p$'s score for method $m$ on case $i$ and criterion $k$. We first average across participants within each case,
\begin{equation}
    \bar s_{i,m,k}=\frac{1}{|\mathcal P_i|}\sum_{p\in\mathcal P_i}s_{p,i,m,k},
    \qquad
    W_{m,k}=\frac{100}{N}\sum_{i=1}^{N}
    \mathbf{1}\!\left[m=\operatorname*{arg\,max}_{m'}\bar s_{i,m',k}\right],
    \label{eq:user_study_win_rate}
\end{equation}
where $\mathcal P_i$ contains the participants rating case $i$, and $N$ is the number of evaluated cases. The reported first-place share $W_{m,k}$ is thus a case-level win rate, not a participant-level vote share or a normalized mean rating. No ties for the highest case-level mean were observed, so each case contributes exactly one winner per criterion and the four shares sum to $100\%$. Figure~\ref{fig:user_study} gives the resulting percentages, rounded to one decimal place. Only aggregate win rates are disclosed.

\section{Additional Qualitative Examples of the Transport Effect}
\label{app:transport_extra}
\begin{figure}[tbp]
  \centering
  \includegraphics[width=0.62\textwidth]{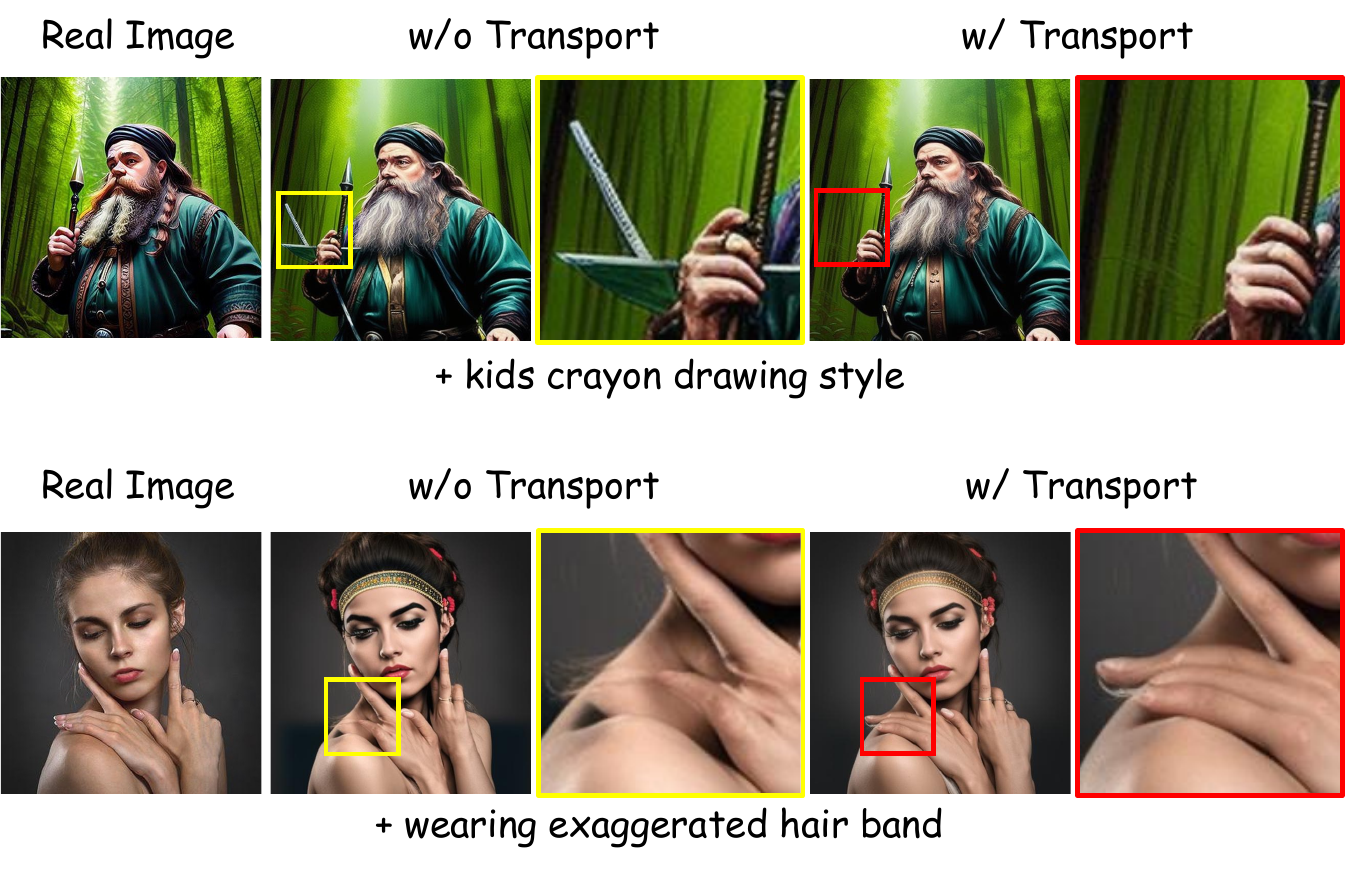}
  \caption{Additional transport ablations complementing Figure~\ref{fig:transport_visual}. From left to right: source, result without transport and its crop, then the ``w/ Transport'' result and its crop. The highlighted regions show an extra shaft and altered hand--shoulder geometry without transport, reduced with the ``w/ Transport'' result.}
  \label{fig:transport_extra}
\end{figure}

The two cases use PIE-Bench and SD-Turbo under the default single-evaluation setting. The crops highlight an unintended extra shaft in the stylization example and altered hand--shoulder geometry in the hair-band example. The ``w/ Transport'' result reduces these visible artifacts in the shown cases.
\par\medskip

\section{Additional Qualitative Results}
\label{app:extra_qualitative}

\begin{figure}[htbp]
  \centering
  \includegraphics[width=\linewidth]{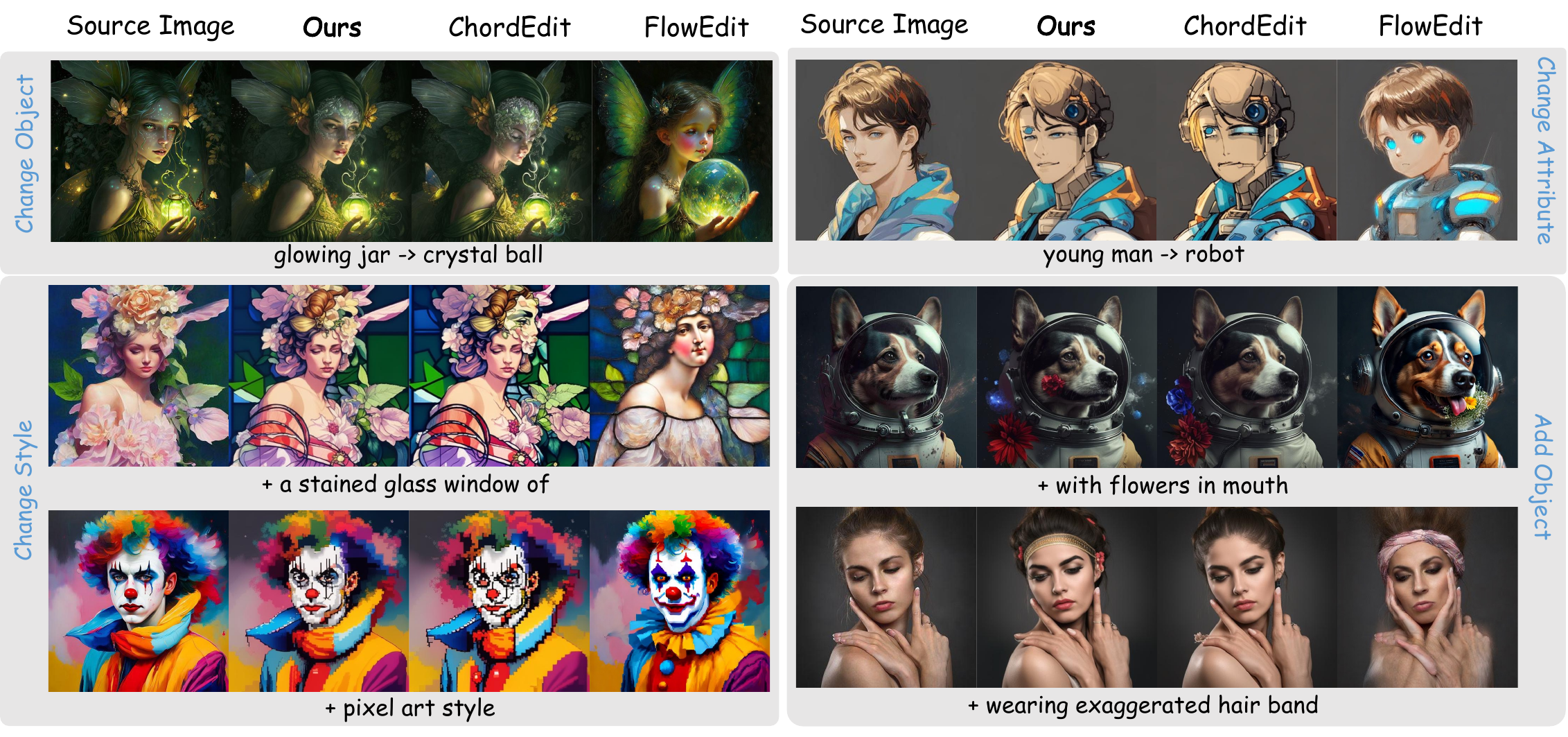}
  \caption{Six qualitative comparisons on PIE-Bench, with source, ECT (ours), ChordEdit, and FlowEdit in each case. Left: object replacement and two style edits. Right: a robot transformation and two object additions. The examples show differences in edit extent and source preservation, including incomplete spatial placement in the flowers-in-mouth case.}
  \label{fig:extra_qualitative}
\end{figure}

Figure~\ref{fig:extra_qualitative} compares ECT, ChordEdit, and FlowEdit on six PIE-Bench cases. The left group contains a glowing-jar-to-crystal-ball edit, stained-glass stylization, and pixel-art stylization. The right group contains a young-man-to-robot edit, flowers-in-mouth addition, and exaggerated-hair-band addition. Each case orders the source, ECT, ChordEdit, and FlowEdit from left to right.

ECT and ChordEdit produce explicit stained-glass boundaries and pixel blocks while retaining much of the source composition. In the robot example, ECT retains more facial appearance while ChordEdit applies a more extensive mechanical transformation. The hair-band case shows the requested accessory in ECT's output together with a broadly preserved hand pose. In the flower example, ECT places flowers near the dog's face and suit, with spatial placement remaining imperfect. These cases illustrate local edits and source preservation with a single update.

\section{Additional Experiments: Seed Stability, Extra-NFE Refinement, and Edit Strength}
\label{app:extra_ablations}

{\color{black}
These PIE-Bench SD-Turbo experiments use a one-update core followed by refinement at $t_{\rm prox}=0.3$ for the seed and $\lambda$ sweeps. Their alignment column uses conventional CLIP with the ViT-B/32 encoder. The refinement-level sweep uses a separately evaluated sample set.
}

\subsection{Stability Across Random Seeds}
\label{app:seed_stability}

ECT draws a single Gaussian sample $\epsilon$ to form the shared noisy state $z_t$ in Equation~\ref{eq:vp_forward}, and this is the only stochastic input to the entire editor: the transport in Equation~\ref{eq:nominal_transport} and the correction in Equation~\ref{eq:ect_update} are deterministic functions of the resulting predictions. A concern for single-evaluation editing is dependence on a favorable noise draw. Table~\ref{tab:seed_stability} evaluates eleven seeds spanning four orders of magnitude on all PIE-Bench cases, with other settings fixed.

\begin{table}[htbp]
\caption{Seed stability on PIE-Bench (SD-Turbo). Eleven independent seeds, with extra-NFE refinement enabled and all other settings fixed. The standard deviations across seeds are about an order of magnitude smaller than the ECT--ChordEdit gap on every preservation metric, and the \emph{worst} seed still outperforms matched ChordEdit on all four. The default seed is shaded blue; the reference row matches ChordEdit in Table~\ref{tab:pie_results} under the same refinement budget.}
    \label{tab:seed_stability}
    \centering
    \small
    \setlength{\tabcolsep}{6pt}
    \resizebox{\linewidth}{!}{\begin{tabular}{lccccc}
        \toprule
        & \multicolumn{4}{c}{Consistency (background preservation)} & \\
        \cmidrule(lr){2-5}
        Seed & PSNR $\uparrow$ & MSE $\downarrow$ & LPIPS $\downarrow$ & SSIM $\uparrow$ & CLIP-B/32 $\uparrow$ \\
        \midrule
        1       & 23.428 & 6.641 & 112.832 & 77.825 & 30.23 \\
        20      & 23.598 & 6.435 & 110.950 & 77.834 & 30.43 \\
        64      & 23.550 & 6.447 & 111.065 & 77.816 & 30.28 \\
        100     & 23.617 & 6.452 & 111.691 & 77.775 & 30.34 \\
        512     & 23.548 & 6.513 & 112.784 & 77.888 & 30.42 \\
        1024    & 23.680 & 6.354 & 111.004 & 77.940 & 30.34 \\
        1536    & 23.544 & 6.514 & 113.927 & 77.859 & 30.34 \\
        2048    & 23.595 & 6.402 & 112.697 & 77.948 & 30.31 \\
        4096    & 23.503 & 6.630 & 113.417 & 77.833 & 30.28 \\
        100000  & 23.613 & 6.418 & 112.478 & 77.874 & 30.24 \\
        \rowcolor{blue!8} 42 (default) & 23.740 & 6.220 & 109.040 & 78.100 & 30.20 \\
        \midrule
        Mean $\pm$ std & 23.583\,{\tiny$\pm$0.084} & 6.457\,{\tiny$\pm$0.119} & 111.990\,{\tiny$\pm$1.399} & 77.881\,{\tiny$\pm$0.089} & 30.31\,{\tiny$\pm$0.07} \\
        Worst seed & 23.428 & 6.641 & 113.927 & 77.775 & 30.20 \\
        \addlinespace[2pt]
        \textit{Ref.}: ChordEdit (SD-T) & \textit{22.630} & \textit{7.990} & \textit{122.640} & \textit{76.710} & \textit{30.60} \\
        \bottomrule
    \end{tabular}}
\end{table}

\paragraph{Takeaway.}
The seed-induced standard deviation is $0.084$ dB PSNR, $0.119$ MSE, $1.399$ LPIPS, and $0.089$ SSIM, i.e.\ a relative dispersion of $0.36\%$, $1.85\%$, $1.25\%$, and $0.11\%$ of the corresponding mean. These are $9.7$--$15.6\times$ smaller than the improvement ECT obtains over matched ChordEdit on the same metrics ($1.11$ PSNR, $1.77$ MSE, $13.60$ LPIPS, $1.39$ SSIM). Every seed, including the worst, exceeds the matched refinement-enabled ChordEdit reference on all four preservation metrics. CLIP spans $0.23$ points across seeds. Together, these results show consistent background preservation across independently sampled noisy states.

\subsection{Supplementary Refinement with an Extra NFE}
\label{app:tprox}

{\color{black}
Tables~\ref{tab:pie_results}--\ref{tab:snr_results} report both refinement budgets. The refinement-level sweep below evaluates a separate PIE sample set. Its additional prediction updates the latent by re-noising and denoising.
}

The latent $x_0$ supplied to refinement is re-noised and denoised as
\begin{equation}
    z_{t_{\mathrm{prox}}}
    =\alpha_{t_{\mathrm{prox}}}x_0
    +\sigma_{t_{\mathrm{prox}}}\epsilon,
    \qquad
    \hat x_0
    =\frac{z_{t_{\mathrm{prox}}}
    -\sigma_{t_{\mathrm{prox}}}
    \hat\epsilon_\phi(z_{t_{\mathrm{prox}}},t_{\mathrm{prox}})}
    {\alpha_{t_{\mathrm{prox}}}},
    \label{eq:tprox_definition}
\end{equation}
Table~\ref{tab:tprox} evaluates the extra-NFE refinement in Equation~\ref{eq:tprox_definition} using ImgEdit-Judge. The sample set is specific to this sweep. Its schedule factors parameterize the supplementary noise level, while Equation~\ref{eq:closed_form} defines the gain at the main editing time.

\begin{table}[htbp]
\caption{Supplementary extra-NFE refinement on PIE-Bench (SD-Turbo). This is a separate historical cohort from the main-table refinement-enabled rows. Among the tested refinement levels, $t_{\mathrm{prox}}=0.3$ (blue) has the highest CP, VQ, and PC. The ChordEdit reference retains its recorded configuration. Best value per column in bold.}
    \label{tab:tprox}
    \centering
    \small
    \setlength{\tabcolsep}{5pt}
    \resizebox{\linewidth}{!}{\begin{tabular}{lccccccc}
        \toprule
        & \multicolumn{4}{c}{VP schedule} & \multicolumn{3}{c}{ImgEdit-Judge} \\
        \cmidrule(lr){2-5}\cmidrule(lr){6-8}
        Setting & $\alpha_{\mathrm{prox}}$ & $\sigma_{\mathrm{prox}}$ & $\theta_{\mathrm{prox}}/\sigma_{\mathrm{prox}}$ & $\theta_{\mathrm{prox}}/\sigma_{\mathrm{prox}}-\zeta$ & CP $\uparrow$ & VQ $\uparrow$ & PC $\uparrow$ \\
        \midrule
        $t_{\mathrm{prox}}=0.1$ & 0.945634 & 0.325234 & 1.019 & 0.319 & 2.92 & 4.33 & 3.99 \\
        \rowcolor{blue!8} $t_{\mathrm{prox}}=0.3$ & 0.768440 & 0.639921 & 1.085 & 0.385 & \textbf{4.04} & \textbf{4.50} & \textbf{4.18} \\
        $t_{\mathrm{prox}}=0.5$ & 0.525673 & 0.850686 & 1.196 & 0.496 & 3.48 & 4.35 & 3.96 \\
        $t_{\mathrm{prox}}=0.7$ & 0.288320 & 0.957534 & 1.335 & 0.635 & 3.30 & 4.15 & 3.59 \\
        \addlinespace[2pt]
        \textit{Ref.}: ChordEdit (SD-T) & --- & --- & --- & --- & \textit{3.95} & \textit{4.37} & \textit{4.00} \\
        \bottomrule
    \end{tabular}}\\[4pt]
    {\footnotesize Subscript $\mathrm{prox}$ denotes evaluation at $t_{\mathrm{prox}}$, with $\theta_{\mathrm{prox}}=\operatorname{atan2}(\sigma_{\mathrm{prox}},\alpha_{\mathrm{prox}})$ and $\zeta=0.7$. These schedule-derived factors parameterize the added noise-and-denoise prediction.}
\end{table}

\paragraph{Takeaway.}

{\color{black}
Within this separate cohort, $t_{\rm prox}=0.3$ gives CP/VQ/PC of $4.04/4.50/4.18$, the highest values among the tested levels. The ChordEdit row supplies the separately recorded reference. The main-table $+$ variants use 0.3 as their recorded refinement setting.
}

\subsection{Edit-Strength Coefficient \texorpdfstring{$\lambda$}{lambda}}
\label{app:lambda}

Table~\ref{tab:lambda} varies $\lambda\in\{1,\dots,6\}$ with refinement enabled and all other settings fixed. This coefficient scales the full update in Equation~\ref{eq:closed_form}, preserving its direction without another evaluation.

\begin{table}[!htbp]
\caption{Effect of the edit-strength coefficient $\lambda$ on PIE-Bench (SD-Turbo). Preservation degrades monotonically with $\lambda$; CLIP peaks at $\lambda=2$ and then declines. Its $0.50$-point gain costs $2.690$ dB PSNR and $37.636$ scaled LPIPS units. The default $\lambda=1$ includes extra refinement.}
    \label{tab:lambda}
    \centering
    \small
    \setlength{\tabcolsep}{6pt}
    \resizebox{\linewidth}{!}{\begin{tabular}{lccccc}
        \toprule
        & \multicolumn{4}{c}{Consistency (background preservation)} & \\
        \cmidrule(lr){2-5}
        Setting & PSNR $\uparrow$ & MSE $\downarrow$ & LPIPS $\downarrow$ & SSIM $\uparrow$ & CLIP-B/32 $\uparrow$ \\
        \midrule
        \rowcolor{blue!8} $\lambda=1$ (ours) & \textbf{23.740} & \textbf{6.220} & \textbf{109.040} & \textbf{78.100} & 30.20 \\
        $\lambda=2$ & 21.050 & 11.655 & 146.676 & 74.432 & \textbf{30.70} \\
        $\lambda=3$ & 19.084 & 18.150 & 183.486 & 71.209 & 30.60 \\
        $\lambda=4$ & 17.724 & 24.364 & 213.623 & 68.508 & 30.20 \\
        $\lambda=5$ & 16.717 & 30.136 & 239.572 & 66.179 & 29.90 \\
        $\lambda=6$ & 15.933 & 35.463 & 263.881 & 64.077 & 29.70 \\
        \addlinespace[2pt]
        \textit{Ref.}: ChordEdit (SD-T) & \textit{22.630} & \textit{7.990} & \textit{122.640} & \textit{76.710} & \textit{30.60} \\
        \bottomrule
    \end{tabular}}
\end{table}

\paragraph{Takeaway.}

{\color{black}
With refinement enabled, $\lambda=1$ favors preservation and $\lambda=2$ maximizes reported CLIP. From $\lambda=1$ to 6, PSNR drops 7.807 dB and scaled LPIPS increases 154.841. This sweep changes strength along a fixed axis and illustrates the resulting trade-off.
}

\setcounter{topnumber}{4}
\setcounter{totalnumber}{6}
\section{Metric Cohorts and Automatic Regions}
\label{app:region_protocol}

\paragraph{CLIP and CLIP-BPB.}
CLIP is the conventional full-image/target-text cosine similarity. CLIP-BPB uses the same encoders and adds a background-preservation weight to each image before averaging. For a common evaluated set $V$, the displayed scores are
\begin{equation}
\begin{aligned}
c_i&=100\cos\!\big(f_I(I_i^{\rm edit}),f_T(e_i)\big),\\
\operatorname{CLIP}(V)&=\frac{1}{|V|}\sum_{i\in V}c_i,\\
\operatorname{CLIP\! -\! BPB}(V)&=\frac{1}{|V|}\sum_{i\in V}c_i\exp(-\beta d_i),\qquad\beta=1.8.
\end{aligned}
\label{eq:clip_bpb_comparison}
\end{equation}
Here $d_i$ is unscaled SqueezeNet LPIPS between the source and edited backgrounds. The exponential factor decreases as background change increases, while the CLIP term retains target-text compatibility. Setting $\beta=0$ gives conventional CLIP on the same evaluated cases. Both scores use ViT-B/32 and the cosine $\times100$ scale; the weight $1.8$ is fixed across the two benchmarks.

\textbf{PIE.} Official edit-region annotations are decoded at $512\times512$. We exclude the one-pixel image boundary from background scoring. The shared valid-background set contains 556 of the 700 cases; every method uses these same valid case IDs for BPB.

\textbf{SNR.} We estimate the requested edit region from the source image with CLIPSeg~\citep{luddecke2022clipseg}. Changed objects or attributes in the source and target descriptions provide the text queries; insertion edits use the nearest relevant source object. Query masks are merged at probability threshold $0.5$, closed with a $9\times9$ neighborhood and expanded by $33\times33$ dilation. The edited area must occupy at least $0.5\%$ of the image and leave at least $5\%$ background. Sixty-five cases yield valid localized regions. For 15 global edits or cases with insufficient localized background, we measure LPIPS over the image interior. Every method uses the same masks and fallback cases. BPB-Auto averages all 80 cases under this protocol.

\begin{table}[t]

\caption{CLIP-BPB weight sensitivity on 556 shared PIE cases. Entries are ECT minus ChordEdit. Setting $\beta=0$ gives CLIP on the same cases, and $\beta=1.8$ is used for both benchmarks.}
\label{tab:bpb_sensitivity}
\centering\small
\begin{tabular}{lrrrrr}
\toprule
Original backbone / refinement & $\beta=0$ & 0.5 & 1.0 & 1.8 & 3.0 \\
\midrule
SD-T$^0$ & -0.1774 & +0.1508 & +0.4453 & +0.8547 & +1.3473 \\
SD-T$^+$ & -0.3922 & -0.1828 & -0.0027 & +0.2338 & +0.4930 \\
SB-v2$^0$ & -0.1509 & +0.1103 & +0.3485 & +0.6863 & +1.1060 \\
SB-v2$^+$ & -0.3184 & -0.1695 & -0.0408 & +0.1292 & +0.3173 \\
\bottomrule
\end{tabular}
\end{table}
Table~\ref{tab:bpb_sensitivity} varies the preservation coefficient on the same 556 PIE cases. The selected coefficient $1.8$ gives ECT gains of $0.1292$--$0.8547$ across the four original-backbone/refinement pairs. The coefficient controls how strongly background changes reduce the score. Selected after PIE method comparisons, the coefficient remains fixed across SNR evaluations.

\section{Evaluation Records and Hardware}
\label{app:provenance}
Table~\ref{tab:pie_results} contains 23 retained configurations and 10 additional flow/native configurations. Table~\ref{tab:snr_results} contains the original 17 configurations and four unrefined ChordEdit/ECT configurations. Preservation, judge and resource columns follow the benchmark records; BPB is computed from reproduced outputs for these configurations using the valid sets above. Thus each column retains its specified measurement cohort. Avg.\ rank equally averages the 12 per-column ranks, calculated from stored unrounded aggregates using per-metric average ranks for ties.

Original resource profiles use NVIDIA H20. The added PIE flow/native profiles and scalar controls use NVIDIA A800-SXM4-80GB. Main-table superscript A marks the latter profiles. Paired speed and resource savings compare the same backbone, device and refinement setting; overall Avg.\ rank summarizes the complete table, including both device groups.

\section{Flow-Backbone Evaluation Protocol}
\label{app:flow_derivation}
\paragraph{Generalized VP coordinates.}
FLUX.1-Schnell~\citep{bfl2024fluxschnell} and SD3.5~\citep{stabilityai2024sd35} use the linear data-to-noise path of rectified flow~\citep{liu2023rectifiedflow,esser2024sd3}. For its interpolation coefficient $\tau\in(0,1)$, write
\begin{equation}
z_\tau=(1-\tau)x_0+\tau\epsilon,\qquad
r_\tau=\sqrt{(1-\tau)^2+\tau^2},\qquad
\bar z_\tau=\frac{z_\tau}{r_\tau}.
\label{eq:flow_native_normalization}
\end{equation}
Define $\alpha_\tau=(1-\tau)/r_\tau$ and $\sigma_\tau=\tau/r_\tau$. Direct substitution gives
\begin{equation}
\bar z_\tau=\alpha_\tau x_0+\sigma_\tau\epsilon,\qquad
\alpha_\tau^2+\sigma_\tau^2=1,\qquad
\theta_\tau=\operatorname{atan2}(\tau,1-\tau).
\label{eq:flow_generalized_vp}
\end{equation}
Thus the normalized path has the VP form of Equation~\ref{eq:vp_forward}, with $\alpha_\tau=\cos\theta_\tau$ and $\sigma_\tau=\sin\theta_\tau$ as in Equation~\ref{eq:angular_parameterization}. The conversion uses state scaling and angular reparameterization of the linear path.

\paragraph{Native velocity and endpoint conversion.}
With velocity oriented from data to noise, $v=\epsilon-x_0$, a native conditional prediction $\hat v^c=\hat v(z_\tau,\tau,c)$ supplies both endpoints:
\begin{equation}
\hat x_0^c=z_\tau-\tau\hat v^c,\qquad
\hat\epsilon^c=z_\tau+(1-\tau)\hat v^c.
\label{eq:flow_endpoint_conversion}
\end{equation}
Differentiating $r_\tau$ and $\theta_\tau$ gives $dr_\tau/d\tau=(2\tau-1)/r_\tau$ and $d\theta_\tau/d\tau=1/r_\tau^2$. The chain rule then converts the predicted native velocity into the tangent of the normalized angular path:
\begin{equation}
\begin{aligned}
\hat v_\theta^c
&=\frac{d\tau}{d\theta_\tau}
\left(\frac{\hat v^c}{r_\tau}-\frac{z_\tau}{r_\tau^2}\frac{dr_\tau}{d\tau}\right)\\
&=\frac{1-2\tau}{r_\tau}z_\tau+r_\tau\hat v^c
=\alpha_\tau\hat\epsilon^c-\sigma_\tau\hat x_0^c.
\end{aligned}
\label{eq:flow_angular_conversion}
\end{equation}
This recovers Equation~\ref{eq:endpoint_velocity} directly from the model's native flow-velocity output.

\paragraph{Shared edit axis and calibrated update.}
Let $\Delta\hat v=\hat v^e-\hat v^s$ at the shared native state. The state-dependent terms cancel between conditions, yielding
\begin{equation}
\begin{aligned}
d_\tau&=\hat x_0^e-\hat x_0^s=-\tau\Delta\hat v,\\
g_\tau&=\hat v_\theta^e-\hat v_\theta^s
=r_\tau\Delta\hat v=-\frac{d_\tau}{\sigma_\tau}.
\end{aligned}
\label{eq:flow_shared_axis}
\end{equation}
The collinearity of Equation~\ref{eq:collinearity} therefore holds in normalized VP coordinates. The nominal transport and endpoint-calibrated displacement follow the same closed-form rule:
\begin{equation}
\widehat a_\tau=-\theta_\tau g_\tau,\qquad
u_\tau=\lambda(\widehat a_\tau-\zeta d_\tau)
=\lambda\left(\frac{\theta_\tau}{\sigma_\tau}-\zeta\right)d_\tau.
\label{eq:flow_ect_update}
\end{equation}
Model queries remain at the native state $z_\tau$; state normalization and output conversion are analytic and reuse the same source--target predictions. For shifted schedules, $\tau$ is the actual interpolation coefficient used to construct the noisy latent. These converted quantities drive the PIE F-S/SD3.5 experiments with the same transport-and-calibration construction.

\section{Full Scalar Controls and Encoder Evaluation}
\label{app:cross_time_reproduction}

\begin{table}[t]

\caption{Validation-calibrated scalar control on SD-Turbo. A 140-case split selects one scalar at $t=0.75$; the scalar is frozen before the disjoint 560-case test. Deltas are ECT minus the frozen scalar on the 438 common valid test backgrounds.}
\label{tab:calibrated_scalar}
\centering\small
\begin{tabular}{rrrrr}
\toprule
$t$ & ECT $\kappa(t)$ & $\Delta$PSNR & $\Delta$LPIPS$\times1000$ & $\Delta$BPB \\
\midrule
0.30 & 0.385127 & +0.137 & -0.448 & -0.0306 \\
0.45 & 0.464331 & +0.495 & -1.911 & +0.0211 \\
0.60 & 0.563594 & +0.654 & -4.059 & +0.1031 \\
0.75 & 0.669884 & 0.000 & 0.000 & 0.0000 \\
0.90 & 0.761418 & -0.918 & +14.887 & -0.5161 \\
\bottomrule
\end{tabular}
\par\vspace{3pt}\footnotesize At $t=0.75$, both methods obtain PSNR $25.749$, LPIPS $0.066011$, and BPB $25.4605$; selected $s=0.669884$.
\end{table}
\textbf{Validation-calibrated scalar.} At fixed time, Equation~\ref{eq:closed_form} selects the coefficient of a shared endpoint direction. We use 140 PIE cases to select a scalar and 560 disjoint cases to test it, with refinement disabled. Candidates share the noisy state, predictions and direction. The grid $\{0.1,0.2,\ldots,1.5\}$ also includes the exact ECT gain. Selection maximizes validation text similarity subject to mean background LPIPS at most that of ECT. At $t=0.75$, the selected scalar $0.669884$ equals the analytic gain: both obtain test PSNR $25.749$, LPIPS $0.066011$, and BPB $25.4605$. ECT therefore reaches this validation-selected operating point directly from the schedule.

\textbf{Transfer across time.} Table~\ref{tab:calibrated_scalar} freezes the selected scalar while varying $t$. At $t=0.60$, ECT improves PSNR by $0.654$ dB and BPB by $0.1031$. At $t=0.90$, its larger coefficient increases background change, with BPB lower by $0.5161$. The schedule-dependent coefficient controls the strength of the endpoint displacement across noise levels.

The full cross-time experiment evaluates all 700 cases at five times using four rules: raw residual, fixed $0.7$, angular transport, and ECT. All rules at each time share the noisy state and model predictions. This produces 14,000 outputs, evaluated with ViT-B/32.
\begin{table}[t]

\caption{Fresh 700-case SD-Turbo reproduction. CLIP-B/32 uses all 700 cases; preservation and BPB use 556 common valid backgrounds. LPIPS is scaled by 1000. No extra refinement is used.}
\label{tab:cross_time_full}
\centering\footnotesize
\setlength{\tabcolsep}{5pt}
\begin{tabular}{rlrrrrr}
\toprule
$t$ & Variant & $s(t)$ & PSNR & LPIPS & CLIP-B/32 & BPB \\
\midrule
0.30 & Raw & 1.000000 & 31.596 & 29.742 & 28.4064 & 26.7959 \\
0.30 & Fixed 0.7 & 0.700000 & 31.799 & 29.055 & 28.3606 & 26.7776 \\
0.30 & Transport & 1.085127 & 31.530 & 29.972 & 28.4186 & 26.7952 \\
0.30 & ECT & 0.385127 & 31.953 & 28.561 & 28.3168 & 26.7558 \\
0.45 & Raw & 1.000000 & 30.100 & 36.198 & 28.5789 & 26.6581 \\
0.45 & Fixed 0.7 & 0.700000 & 30.904 & 32.288 & 28.4683 & 26.7253 \\
0.45 & Transport & 1.164331 & 29.641 & 38.822 & 28.6470 & 26.6063 \\
0.45 & ECT & 0.464331 & 31.440 & 30.089 & 28.3960 & 26.7568 \\
0.60 & Raw & 1.000000 & 26.883 & 57.835 & 28.8079 & 25.8993 \\
0.60 & Fixed 0.7 & 0.700000 & 28.615 & 43.812 & 28.5867 & 26.3194 \\
0.60 & Transport & 1.263594 & 25.534 & 71.863 & 28.9321 & 25.4249 \\
0.60 & ECT & 0.563594 & 29.442 & 38.584 & 28.4973 & 26.4671 \\
0.75 & Raw & 1.000000 & 23.130 & 100.190 & 28.9743 & 24.2550 \\
0.75 & Fixed 0.7 & 0.700000 & 25.471 & 69.425 & 28.7224 & 25.3071 \\
0.75 & Transport & 1.369884 & 20.995 & 135.810 & 28.9876 & 22.8632 \\
0.75 & ECT & 0.669884 & 25.739 & 66.428 & 28.6875 & 25.4060 \\
0.90 & Raw & 1.000000 & 19.840 & 152.769 & 28.9588 & 22.1459 \\
0.90 & Fixed 0.7 & 0.700000 & 22.416 & 105.720 & 28.7577 & 23.7823 \\
0.90 & Transport & 1.461418 & 17.339 & 210.466 & 29.1532 & 20.2414 \\
0.90 & ECT & 0.761418 & 21.808 & 115.779 & 28.8080 & 23.4342 \\
\bottomrule
\end{tabular}
\end{table}
\begin{figure}[!htbp]
\centering
\includegraphics[width=0.94\linewidth]{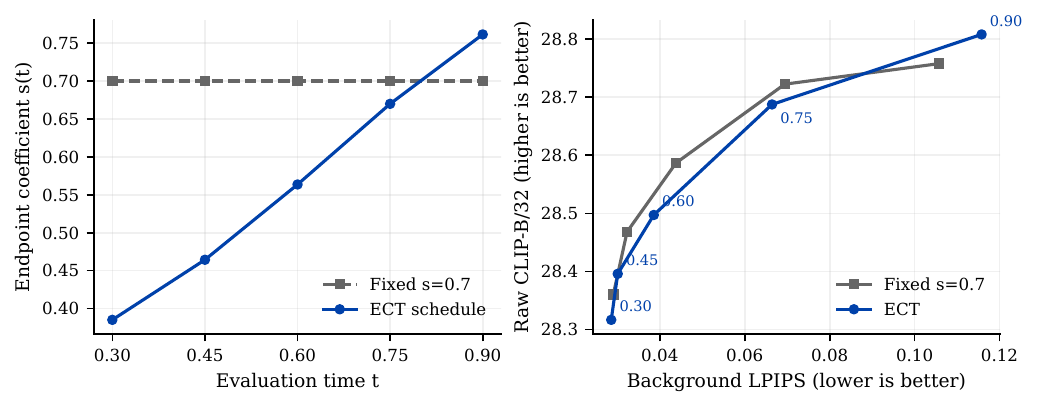}
\caption{Scalar ablation across time. Left: fixed and ECT coefficients. Right: preservation--alignment operating points under ViT-B/32. At $t\le0.75$, ECT's smaller coefficient improves preservation; at $0.90$, its larger coefficient increases raw text similarity and background change.}
\label{fig:cross_time_reproduction}
\end{figure}
At $t=0.60$, ECT improves PSNR from $28.615$ to $29.442$ and BPB from $26.3194$ to $26.4671$ relative to fixed $0.7$. At $t=0.90$, fixed $0.7$ gives PSNR $22.416$ and BPB $23.7823$, versus ECT's $21.808$ and $23.4342$. The complete sweep shows the effect of schedule-dependent edit strength on preservation and alignment.

\begin{table}[t]

\caption{CLIP encoder comparison at $t=0.75$ on 700 saved outputs per rule. Both encoders use the same images and target texts; scores are cosine similarity $\times100$.}
\label{tab:encoder_check}
\centering\small
\begin{tabular}{lrrr}
\toprule
Variant & ViT-B/32 & ViT-L/14 & Difference \\
\midrule
Raw & 28.9743 & 23.8635 & +5.1109 \\
Fixed 0.7 & 28.7224 & 23.1808 & +5.5416 \\
Transport & 28.9876 & 24.4529 & +4.5347 \\
ECT & 28.6875 & 23.0973 & +5.5902 \\
\bottomrule
\end{tabular}
\end{table}
Table~\ref{tab:encoder_check} scores the same 700 outputs per rule with two CLIP encoders. At $t=0.75$, ECT scores $28.6875$ with B/32 and $23.0973$ with L/14. The earlier L/14 scalar run scored $23.072617$, while the original B/32 experiment scored $28.60$ on a separately generated image set. The same-image comparison isolates the encoder's contribution to the absolute score scale. All BPB results use B/32.

\section{\texorpdfstring{Transport and Parameter Ablations}{Transport and Parameter Ablations}}
\label{app:legacy_ablations}

{Table~\ref{tab:ablations} compares core update count, angular transport and refinement on PIE-Bench. Its CLIP column uses ViT-B/32; panels (a,b) include refinement, while panel (c) compares refinement budgets.
}

The full three-panel ablation is reported in Table~\ref{tab:ablations} in the main text.

\begin{figure}[!htb]
\centering

    \includegraphics[width=0.55\linewidth]{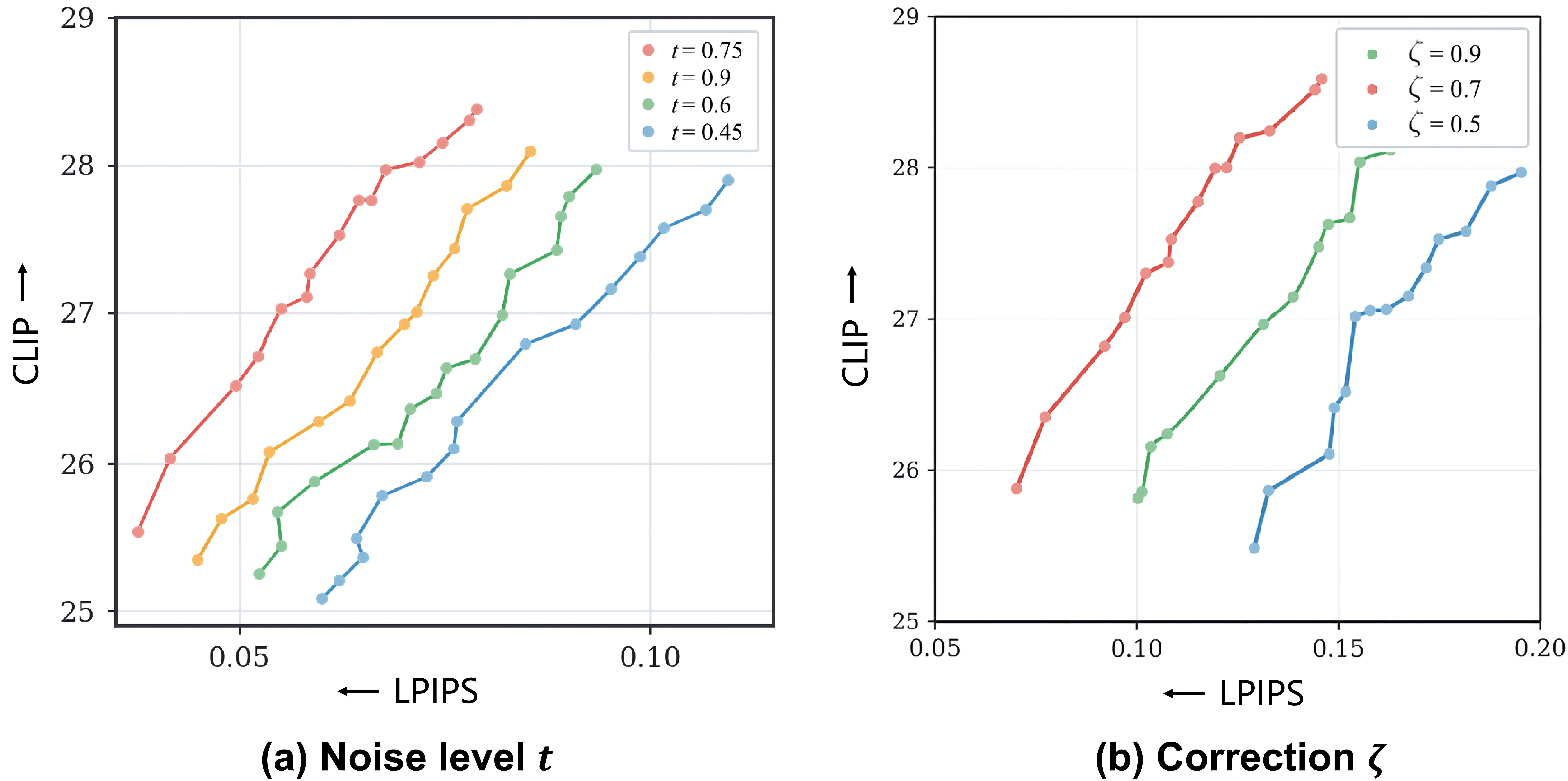}
    
    \caption{CLIP parameter trade-offs on PIE-Bench. Lower LPIPS and higher CLIP are preferable; LPIPS is unscaled. (a) $t=0.75$ gives lower LPIPS at comparable CLIP than the other plotted evaluation times. (b) $\zeta=0.7$ gives the most favorable plotted trade-off.}
    \label{fig:ablation_params}

\end{figure}
\begin{figure}[!htb]
\centering
    \includegraphics[width=0.55\linewidth]{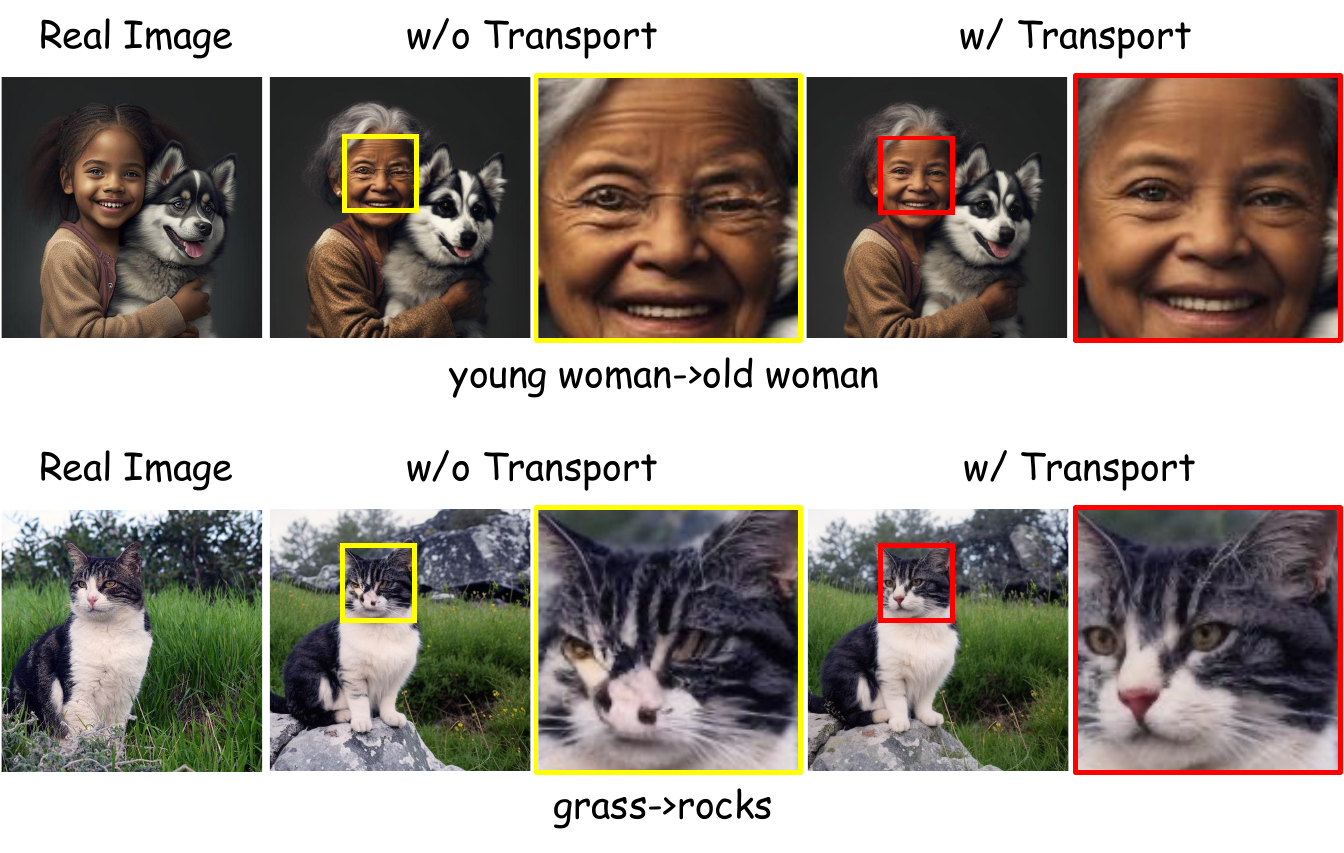}
    
    \caption{Qualitative effects of angular transport on two representative
    cases. From left to right: source, result without transport and its
    enlarged crop, then the ``w/ Transport'' result and its enlarged crop.}

    \label{fig:transport_visual}
\end{figure}
\end{document}